\PassOptionsToPackage{numbers,sort&compress}{natbib}
\documentclass{article}
\usepackage[preprint]{neurips_2026}
\usepackage[utf8]{inputenc}
\usepackage[T1]{fontenc}
\usepackage{url}
\usepackage{amsmath,amssymb,amsfonts,amsthm}
\usepackage{booktabs}
\usepackage{graphicx}
\usepackage{float}
\usepackage{xcolor}
\usepackage{hyperref}
\usepackage{multirow}
\usepackage{algorithm}
\usepackage{algorithmic}
\usepackage{enumitem}
\usepackage{subcaption}
\usepackage{array}
\usepackage{colortbl}
\usepackage{tcolorbox}
\usepackage{microtype}
\usepackage{placeins}
\usepackage{makecell}
\usepackage{rotating}
\usepackage{pdflscape}
\usepackage{arydshln}

\tcbuselibrary{skins,breakable}
\newtcolorbox{promptbox}[1][]{%
  colback=gray!5, colframe=gray!50!black, fonttitle=\bfseries\footnotesize,
  title={#1}, boxrule=0.4pt, arc=1.5pt, left=3pt, right=3pt, top=1pt, bottom=1pt,
  fontupper=\footnotesize\ttfamily, breakable}

\hypersetup{
  colorlinks=true, linkcolor=blue!70!black, citecolor=green!40!black, urlcolor=blue!70!black,
  pdftitle={MOTIF: LLM-Defined Material Taxonomies for OOD Property Prediction via Invariant Feature Selection},
  pdfauthor={Gwangseon Jang, Sieun Yun, Dong Ho Lee, Kyong-Ha Lee, Chanuk Lim, Jin Woo Choi, Han Seul Kim}
}

\newcommand{\ours}{\textsc{Motif}}

\newcommand{\etal}{\textit{et al.}}
\newcommand{\tbfont}{\fontsize{7}{9}\selectfont}
\newcommand{\appfont}{\fontsize{8}{10}\selectfont}
\newcommand{\maintabfont}{\fontsize{7}{8.5}\selectfont}

\title{\ours{}: LLM-Defined Material Taxonomies for\\OOD Property Prediction via Invariant Feature Selection}

\author{%
  \begin{minipage}{\dimexpr\textwidth-2\tabcolsep\relax}\centering
  {\normalsize
  Gwangseon Jang$^{1}$ \quad Sieun Yun$^{2}$ \quad Dong Ho Lee$^{2}$ \quad Kyong-Ha Lee$^{1}$ \\
  Chanuk Lim$^{1}$ \quad Jin Woo Choi$^{3,\ast}$ \quad Han Seul Kim$^{2,\ast}$}\\[0.5em]
  {\footnotesize\normalfont
  $^{1}$ Large-scale AI Research Center, Korea Institute of Science and Technology Information (KISTI)\\
  $^{2}$ Department of Advanced Materials Engineering, Chungbuk National University\\
  $^{3}$ Department of Data Information and Physics, Kongju National University}\\[0.4em]
  {\footnotesize\normalfont
  $^{\ast}$ Co-corresponding authors:\\
  Jin Woo Choi (\texttt{jinwoo.choi@kongju.ac.kr}) \quad
  Han Seul Kim (\texttt{hanseul.kim@chungbuk.ac.kr}).}
  \end{minipage}%
}

\begin{document}
\maketitle

\begin{abstract}
Out-of-distribution (OOD) generalization remains a central challenge in materials property prediction, where models must extrapolate to novel chemistries beyond the training distribution. While large language models (LLMs) have demonstrated strong scientific reasoning, their direct use for quantitative prediction remains limited. To address this, we systematically evaluate multiple roles for LLM assistance, finding that only taxonomy-based grouping consistently improves OOD generalization over a strong kNN+GBT baseline. Building on this, we here propose \ours{}, a framework that leverages LLMs not as predictors, but as generators of candidate taxonomies that partition materials into chemically meaningful environments. These taxonomies are evaluated via training-only pseudo-OOD validation, and only those that improve generalization are retained. Invariant feature selection is then applied within each accepted taxonomy, followed by gated aggregation across taxonomies. We evaluate on 2D halide perovskites, where bandgap prediction must generalize across distinct organic spacer families. This poses a realistic OOD challenge due to the strong influence of spacer chemistry on electronic structure. On this benchmark,  \ours{} reduces mean OOD MAE from 0.122 to 0.111 eV. Under a frozen dev/test protocol, the deployable configuration further improves errors from 0.123 to 0.117 eV (electronic bandgap) and from 0.181 to 0.173 eV (optical bandgap). Beyond this primary task, we validate transferability across additional materials datasets, including 3D perovskite solar cell efficiencies and Matbench benchmarks with structure-based splits. Across domains,  \ours{} consistently improves robustness under distribution shift, outperforming expert-defined groupings, rule-based heuristics, and clustering baselines, while remaining competitive with strong classical selection methods. These results demonstrate that LLMs are most effective not as predictors, but as tools for discovering structure in data. This enables improved generalization in materials science tasks under realistic OOD conditions. 
\end{abstract}

\section{Introduction}
\label{sec:intro}

In materials science, predictive models are often applied to compounds with novel chemistries outside the training distribution. This makes out-of-distribution (OOD) generalization a fundamental requirement. While large language models (LLMs) have demonstrated strong capabilities across scientific domains, their most reliable role in materials property prediction has not been systematically established.~\cite{bran2024augmenting,boiko2023emergent,jablonka2024leveraging}.
Most prior work treats LLMs as direct predictors, using them to generate or refine numerical property values, often with additional techniques such as retrieval, reasoning traces, or self-correction~\cite{jablonka2024leveraging,rubungo2025llmprop,requeima2024llm,wei2022chain,shinn2023reflexion,yao2023tree,zhou2024language,dhuliawala2024chain}. However, under leave-one-spacer-family-out shift, each test fold contains an unseen organic spacer superfamily absent from training, so the model must extrapolate to new spacer chemistries with distinct structure–property relationships rather than interpolate within a shared family. In such settings, even weak spurious correlations can lead to misleading improvements, making apparent performance gains unreliable. This raises a more fundamental question: not whether an LLM can predict the value, but where in the prediction pipeline its knowledge can be reliably used to improve OOD materials prediction. 

To investigate this question, we study leave-one-spacer-family-out prediction for two-dimensional (2D) metal-halide perovskites~\cite{marchenko2020database}.
Each test fold contains an unseen spacer superfamily, requiring generalization beyond local interpolation and reflecting realistic scenarios where models must predict properties for entirely new chemistries.
Here, the organic spacer strongly modulates the electronic structure, so different spacer families induce distinct structure--property relationships.

We evaluate six roles for LLM assistance, including direct prediction, retrieval-augmented prediction, LLM-generated features, LLM-guided sample selection, priors or constraints, and chemical taxonomy annotation, i.e., the use of LLMs to define meaningful groupings of materials. 
Among these, only offline taxonomy annotation consistently improves over a strong k-nearest neighbors (kNN) and gradient boosting trees (GBT) baseline.
In contrast, approaches that directly incorporate LLM-generated numerical values into the predictor yield at most marginal gains, improving the baseline by less than 2\% across 19 published methods and over 65 in-house variants. 
These results suggest that the most reliable use of LLMs in this setting is not as predictors, but as offline chemistry annotators that propose chemically meaningful groupings, which can then be validated using classical statistical methods. 

This observation highlights a key bottleneck in applying invariant learning to scientific data: the lack of meaningful ways to partition the data into groups. 
Invariant learning aims to identify relationships between features and targets that remain stable across such groups~\cite{arjovsky2019invariant,peters2016causal}. However, in real-world datasets, these group structures are rarely explicitly available. Random partitions are often uninformative, while expert-defined groupings are costly and typically too coarse. 
We therefore introduce \ours{} (\textbf{M}aterials \textbf{O}OD \textbf{T}axonomy-\textbf{I}nvariant \textbf{F}ramework), which uses an LLM once, offline, to generate chemistry-aware taxonomies of spacer cations. These taxonomies capture variations in spacer chemistry that influence electronic structure and bandgap formation, thereby defining meaningful groupings of the data. Each taxonomy defines a candidate environment. For each environment, \ours{} selects features whose relationships with the target remain stable across chemically distinct groups using a Chow structural-break test~\cite{chow1960tests}. It then trains predictors using only these invariant features. It then deploys taxonomy-conditioned models only when supported by training-only pseudo-OOD validation. 
Notably, the LLM is used only offline, does not produce numerical property estimates, and is never queried at inference time. 
For the main six-fold analysis, we use a fully automatic gated-median readout. For frozen-protocol and transfer evaluations, we adopt a more conservative safe-blend deployment rule to ensure robust performance under stricter evaluation settings. 

On the six-fold 2D halide perovskite benchmark, \ours{} reduces mean OOD MAE from 0.122 to 0.111~eV.
Under a stricter frozen dev/test protocol, the primary deployable configuration improves bandgap from 0.123 to 0.117~eV ($p{=}0.002$) and optical bandgap from 0.181 to 0.173~eV. It outperforms expert-family, rule-based, and clustering-based methods, while remaining competitive with strong classical feature selection approaches. The same pipeline also improves four additional OOD benchmarks across diverse materials families and target properties.
Taken together, these results suggest that, in  high-precision OOD materials settings, LLMs are more effective for proposing auditable environment hypotheses than for producing numerical property estimates. 

Our contributions are threefold:
\begin{enumerate}[nosep, leftmargin=*]
  \item We recast LLM use in  high-precision scientific OOD prediction as an environment-discovery problem for invariant learning rather than a direct numerical regression problem, and we evaluate that framing across 19 published LLM methods and 65{+} in-house variants.
  
  \item We introduce \ours{}, which converts LLM-generated chemical taxonomies into candidate environments, selects invariant features using Chow-based tests, and deploys taxonomy-conditioned predictors only when supported by training-only pseudo-OOD validation.
  
  \item We provide empirical evidence across a six-fold perovskite benchmark, a frozen dev/test protocol, and four transfer benchmarks that this environment-annotation role is more reliable than direct numerical use of LLMs for  high-precision OOD materials prediction.
\end{enumerate}

\section{Related Work}
\label{sec:related}

\textbf{LLMs for scientific prediction.}
LLMs have been applied to synthesis planning~\cite{bran2024augmenting}, autonomous experimentation~\cite{boiko2023emergent}, and property prediction~\cite{jablonka2024leveraging, rubungo2025llmprop, requeima2024llm}.
Most prior work treats LLMs as predictors, editors, or feature generators.  In contrast, we systematically compare these roles and demonstrate that, for  high-precision OOD materials prediction, offline annotation of chemically meaningful groupings is consistently more reliable than direct numerical use of LLMs. 

\textbf{Invariant learning.}
Invariant risk minimization (IRM) and related methods seek predictors whose relationships to the target remain stable across environments ~\cite{arjovsky2019invariant, gulrajani2021in, peters2016causal}, but typically assume that such environments are given.
In contrast, we generate candidate environments automatically using LLM-derived taxonomies, and combine them with classical structural-break testing and conservative model selection. 

\textbf{OOD materials benchmarks.}
Prior work on materials OOD evaluation includes Matbench~\cite{dunn2020benchmarking}, structure-based splits~\cite{kim2024structure}, and transductive methods~\cite{segal2025known}.
In contrast, we adopt leave-one-family-out splits that explicitly test generalization to genuinely novel chemical families. 


\begin{figure}[!t]
  \centering
  \includegraphics[width=\textwidth, height=5cm, keepaspectratio]{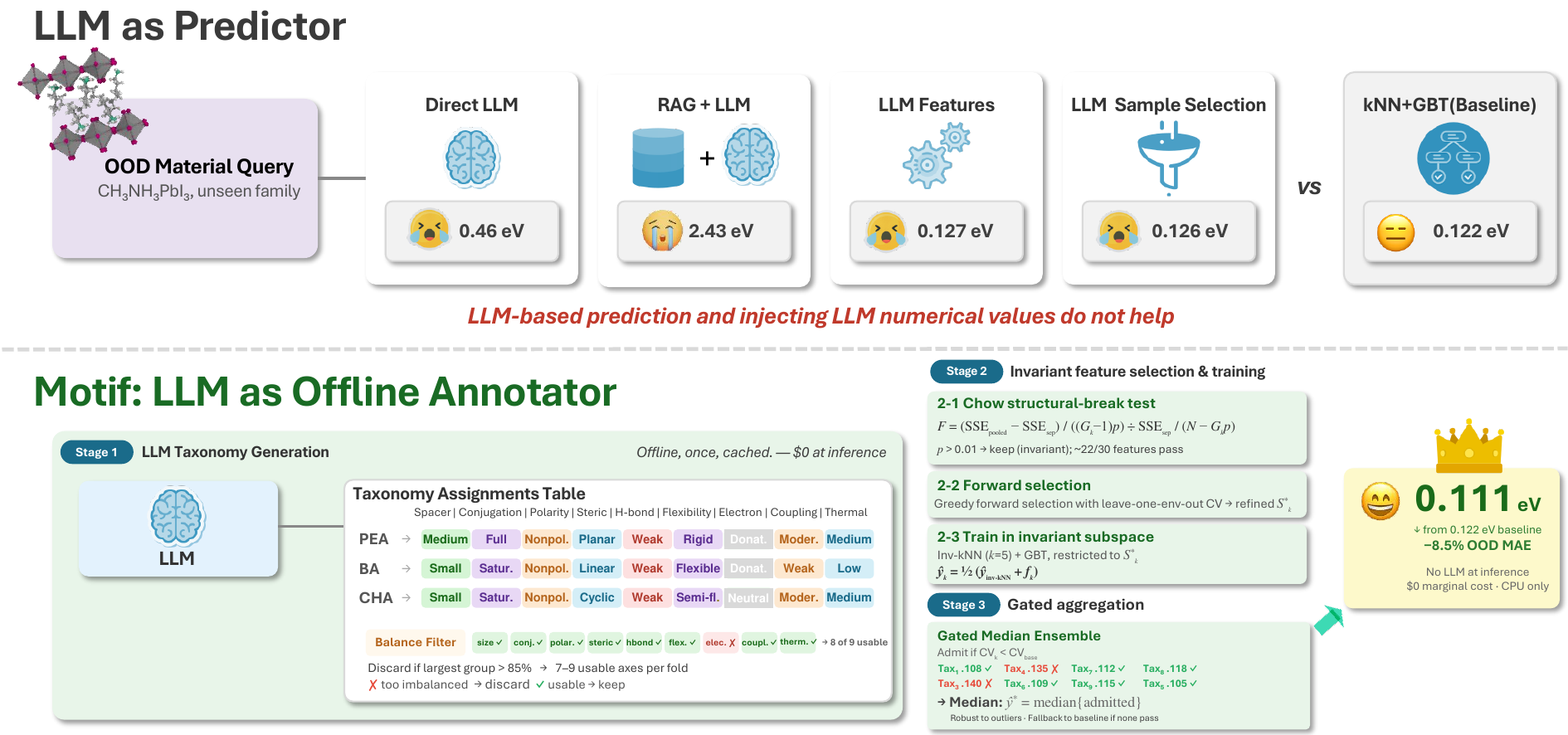}
  \caption{\ours{} pipeline. \textbf{Stage 1}: an LLM classifies each spacer along multiple chemical axes once, offline. \textbf{Stage 2}: within each taxonomy, a Chow test filters features whose relationship to the target is stable across groups, and kNN/GBT are trained only in that invariant subspace. \textbf{Stage 3}: a pseudo-OOD gate keeps only taxonomies that beat the baseline, and gated aggregation combines the surviving models. No LLM call is made at inference time.}
  \label{fig:overview}
\end{figure}

\section{Method}
\label{sec:method}

\subsection{Overview}

Figure~\ref{fig:overview} summarizes the proposed pipeline, which addresses the missing-environment problem by converting environment discovery into a generate--validate--select procedure. 
\ours{} first uses an LLM to generate $K$ candidate chemical taxonomies offline, each representing a hypothesis for grouping the data into environments. 
Each taxonomy is then treated as a candidate environment for invariant feature selection and predictor training. 
We use a training-only pseudo-OOD criterion to filter candidate taxonomies, retaining only those that improve over the baseline and falling back to the baseline otherwise.
The design is deliberately conservative: the LLM proposes candidate structures, while statistical tests and validation determine which ones are reliable.

\subsection{Stage 1: LLM-defined chemical taxonomies}
\label{sec:taxonomy}

We prompt an LLM once, offline, to classify each unique spacer cation along $K=9$ chemically motivated axes: spacer size, conjugation, polarity, steric profile, H-bonding, flexibility, electron effect, interlayer coupling, and thermal stability.
Each axis induces a 2--4 way partition over the spacers, which we interpret as a candidate environment definition.
For the primary 2D perovskite benchmark, this yield annotations for 313 unique spacers; external datasets reuse the same annotation idea with dataset-specific entity counts.
The resulting taxonimies are discrete, auditable, and cached across all folds, making this step a one-time offline preprocessing.
Because the total number of annotations depends on the benchmark and model backbone, we report concrete annotation counts and costs in Section~\ref{sec:cross_benchmark} and Appendix~\ref{app:model_detail}, rather than treating them as universal method constants.

Not all candidate environments are suitable for invariance analysis.
For invariance testing, an environment split must be reasonably balanced; otherwise the small groups produce unstable estimates and the large group dominates the test.
We therefore discard taxonomies whose largest group exceeds 85\% of the available training samples, resulting in 7--9 usable taxonomies per fold.
Appendix~\ref{app:balance} shows that pilot taxonomy banks based on bulk-material notions (e.g. confinement and stability) failed because they produced highly imbalanced (degenerate or nearly degenerate) partitions, making them unsuitable for reliable invariance testing.

\subsection{Stage 2: invariant feature selection}
\label{sec:invariance}

For taxonomy $k$, let $e_k(i) \in \{1, \ldots, G_k\}$ denote the group assignment of training sample $i$. We treat these groups as environments and seek features whose relationship with the target remains stable across them, with the goal of removing environment-specific shortcuts that may not generalize under distribution shift.
For each feature $j$, we test whether its linear relationship with the target is stable across groups using the Chow structural-break test~\cite{chow1960tests}.
This is a conservative \emph{screening} step rather than a modeling assumption: features that exhibit strong environment-specific slope changes are excluded before training nonlinear predictors. 
Let $\mathrm{SSE}_{\mathrm{pooled}}$ be the residual sum of squares from a single regression fit to all training samples, and let $\mathrm{SSE}_{\mathrm{sep}} = \sum_{g=1}^{G_k} \mathrm{SSE}_g$ be the sum from separate regressions within each group.
The test statistic is
\begin{equation}
F = \frac{(\mathrm{SSE}_{\mathrm{pooled}} - \mathrm{SSE}_{\mathrm{sep}}) / ((G_k - 1)p)}{\mathrm{SSE}_{\mathrm{sep}} / (N - G_k p)},
\end{equation}
where $p$ is the number of regression parameters.
We call feature $j$ \emph{invariant} under taxonomy $k$ when the Chow $p$-value exceeds $\alpha = 0.01$.
This threshold is intentionally strict: allowing too many borderline features reintroduces environment-specific correlations that improve in-distribution performance but degrade under family shift.
Appendix~\ref{app:alpha} shows that both stricter and looser thresholds are worse on average.

Given the invariant set $\mathcal{S}_k$, we run a small greedy forward-selection step with leave-one-environment-out validation to obtain a refined subset $\mathcal{S}_k^*$ and a GBT regressor $f_k$.
A practical point matters here: \emph{kNN search must be restricted to the same invariant subspace}. Using the full feature set allows OOD-irrelevant spacer indicators to dominate distance calculations, particularly when the held-out family is absent from training. Restricting kNN to $\mathcal{S}_k^*$ ensures consistency between feature selection and neighborhood definition.

We therefore define
\begin{equation}
\hat{y}_{\mathrm{inv\text{-}kNN}}^{(k)}(\mathbf{x}^*) = \sum_{i \in \mathcal{N}_5(\mathbf{x}_{\mathcal{S}_k^*}^*)} \frac{w_i}{\sum_j w_j} y_i,
\qquad
w_i = \frac{1}{\lVert \mathbf{x}_{\mathcal{S}_k^*}^* - \mathbf{x}^{(i)}_{\mathcal{S}_k^*} \rVert + \epsilon}.
\end{equation}

The per-taxonomy predictor is the average of invariant-kNN and GBT:
\begin{equation}
\hat{y}_k = \tfrac{1}{2}\Big(\hat{y}_{\mathrm{inv\text{-}kNN}}^{(k)} + f_k(\mathbf{x}_{\mathcal{S}_k^*})\Big).
\end{equation}

\subsection{Stage 3: gated aggregation}
\label{sec:ensemble}

Because some taxonomies improve generalization while others are actively harmful under distribution shift, \ours{} aggregates predictions selectively rather than averaging across all taxonomies.
For each taxonomy $k$, we compute leave-one-environment-out validation error using a training-only pseudo-OOD criterion on the training set,
\begin{equation}
\mathrm{CV}_k = \frac{1}{|\mathcal{E}_k|} \sum_{g \in \mathcal{E}_k} \mathrm{MAE}(f_k^{(-g)}, \mathcal{D}_g),
\end{equation}
where $f_k^{(-g)}$ is trained without environment $g$ and $\mathcal{E}_k$ contains environments with at least five samples.
Taxonomy $k$ is admitted only when it improves over the baseline, i.e., $\mathrm{CV}_k < \mathrm{CV}_{\mathrm{base}}$. This ``do no harm'' gate ensures that only taxonomies with demonstrated OOD benefit contribute to the final prediction.
The final prediction is the median of all admitted taxonomy predictions,
\begin{equation}
\hat{y}^* = \mathrm{median}\big(\{\hat{y}_k : \mathrm{CV}_k < \mathrm{CV}_{\mathrm{base}}\}\big),
\end{equation}
with fallback to the baseline if no taxonomy passes. The median provides robustness to occasional mis-specified or overconfident taxonomy predictions.

In the main six-fold benchmark, this gated-median rule is the default fully automatic readout because it exposes fold-level routing behavior most clearly.
For frozen-protocol and transfer evaluations, we additionally report a conservative safe-blend readout that mixes the baseline with the best admitted taxonomy, reducing downside risk when routing is uncertain.

\section{Experiments}
\label{sec:experiments}

\subsection{Setup}
\label{sec:setup}

\textbf{Datasets and splits.}
Our primary benchmark is the public 2D Halide Perovskite Database of Marchenko \etal~\cite{marchenko2020database}, using the NMSE portal snapshot accessed on March~21,~2026~\cite{nmse2026db}. After filtering for valid experimental bandgap values, 753 compounds were retained. For OOD evaluation, we use six leave-one-spacer-family-out folds, where each chemically defined spacer-cation family is held out once for testing: short-alkyl, long-alkyl, phenyl, heterocyclic, cycloalkyl, and special-case groups. Across the six folds, every compound appears exactly once in the OOD test set. To assess transfer beyond the primary domain, we further evaluate on Matbench dielectric and 2D exfoliation~\cite{dunn2020benchmarking}, an OQMD ABX$_3$ formation-energy subset~\cite{kirklin2015oqmd}, and a Materials Project ABX$_3$ bandgap subset~\cite{jain2013materials}, using dataset-specific leave-one-group-out OOD splits (6--7 folds; Section~\ref{sec:cross_benchmark}). Detailed fold statistics are provided in Appendix~\ref{app:dataset}.

\textbf{Features and baseline.}
For the 2D perovskite bandgap and optical-bandgap benchmarks, all methods use the same 30-dimensional compositional feature vector, including structural descriptors, B-site and halide composition, spacer-family indicators, and A-site descriptors (Appendix~\ref{app:features}). For the four transfer benchmarks, we keep the same local predictor template---inverse-distance kNN blended with GBT and wrapped by the same taxonomy/invariance pipeline---but use each dataset's native descriptor space instead of the perovskite-specific features. The baseline is the corresponding kNN+GBT model without taxonomy-conditioned invariant filtering.

\textbf{Evaluation protocol.}
We consider three evaluation settings. First, the six-fold leave-one-spacer-family-out bandgap benchmark on 2D halide perovskites serves as the primary setting for fold-wise analysis, method comparison, and comparison against direct-LLM and RAG baselines. Second, we evaluate a stricter frozen dev/test protocol on perovskite bandgap and optical-bandgap tasks, where key hyperparameters ($\alpha{=}0.01$, 85\% balance, gated median) are selected on a frozen dev set (alkyl-short and phenyl), and the remaining four folds are used as a frozen test set (Appendix~\ref{app:frozen_dev}). Third, we test generalization on four additional transfer benchmarks. To prevent information leakage, taxonomy labels are precomputed for all spacers, but only training-set labels are used during model fitting, and zero spacer overlap is verified for each fold (Appendix~\ref{app:leakage}).

\subsection{Main results on the 2D halide perovskite bandgap benchmark}
\label{sec:main_results}

Table~\ref{tab:main} reports results on six-fold leave-one-spacer-family-out OOD benchmark on 2D halide perovskite bandgap prediction.
For this benchmark, we report results using the fully automatic gated-median readout, while later frozen-protocol sections adopt the safe-blend readout for a more conservative evaluation.

\ours{} (gated median) reaches 0.111~eV mean OOD MAE, improving over the 0.122~eV kNN{+}GBT baseline and achieving the best performance on four of the six folds.
Because the baseline is already strong, the 0.011~eV reduction is practically meaningful.
Two post-hoc reference rows  provide additional context for the remaining performance gap. The best fixed single taxonomy attains 0.110~eV, while the oracle (the best taxonomy chosen separately for each fold with access to test labels) reaches 0.100~eV.
This gap highlights the distinction between \emph{taxonomy quality} from \emph{routing quality}, suggesting that the improvements are primarily limited by the selection strategy rather than the taxonomy itself.

\begin{table}[!t]
\centering
\caption{2D halide perovskite bandgap: six-fold leave-one-spacer-family-out OOD benchmark (MAE in eV). \colorbox{green!12}{\ours{}} is our proposed method. kNN, GBT, and their blend are non-LLM baselines; Self-Refine~\cite{madaan2023selfrefine} is the best-performing direct LLM predictor. ``Best single axis'' and ``oracle'' are post-hoc upper bounds, not deployable configurations.}
\label{tab:main}
{\maintabfont
\begin{tabular}{lcccccccc}
\toprule
Method & Alkyl-S & Alkyl-L & Phenyl & Hetero & Cyclo & Special & Mean & $\Delta$ \\
\midrule
kNN (30-dim) & 0.113 & 0.075 & 0.142 & 0.092 & 0.183 & 0.165 & 0.128 & --- \\
GBT (30-dim) & 0.125 & 0.104 & 0.207 & 0.058 & 0.164 & 0.148 & 0.134 & --- \\
kNN+GBT Blend (baseline) & 0.123 & 0.136 & 0.119 & 0.110 & 0.137 & 0.105 & 0.122 & --- \\
\midrule
Self-Refine~\cite{madaan2023selfrefine} (best LLM) & 0.425 & 0.282 & 0.644 & 0.419 & 0.658 & 0.335 & 0.461 & $+278\%$ \\
\midrule
\rowcolor{green!8} \ours{} (gated median) & \textbf{0.128} & \textbf{0.129} & \textbf{0.095} & \textbf{0.099} & \textbf{0.098} & 0.119 & \textbf{0.111} & $\mathbf{-8.5\%}$ \\
\cdashline{1-9}
\ours{} (best single axis; post hoc) & 0.118 & 0.130 & 0.109 & 0.094 & 0.118 & 0.094 & 0.110 & $-9.3\%$ \\
\ours{} (oracle axis; per-fold) & 0.117 & 0.118 & 0.092 & 0.086 & 0.097 & 0.089 & 0.100 & $-17.8\%$ \\
\bottomrule
\end{tabular}}
\vspace{-6pt}
\end{table}

\begin{table}[!t]
  \vspace{-8pt}
  \centering
    \caption{2D halide perovskite bandgap: fold-level OOD MAE matrix. \ours{} beats the kNN{+}GBT blend on Alkyl-L, Phenyl, Hetero, and Cyclo, but not on Alkyl-S or Special.}
  {\maintabfont
  \setlength{\tabcolsep}{4pt}
  \renewcommand{\arraystretch}{1.15}
  \begin{tabular}{lcccccc}
    \toprule
    Method & Alkyl-S & Alkyl-L & Phenyl & Hetero & Cyclo & Special \\
    \midrule
    \ours{} & \cellcolor{yellow!20}0.128 & \cellcolor{yellow!20}0.129 & \cellcolor{green!25}0.095 & \cellcolor{green!20}0.099 & \cellcolor{green!25}0.098 & \cellcolor{yellow!25}0.119 \\
    kNN{+}GBT Blend & \cellcolor{green!20}0.123 & \cellcolor{orange!25}0.136 & \cellcolor{yellow!25}0.119 & \cellcolor{orange!20}0.110 & \cellcolor{orange!25}0.137 & \cellcolor{green!20}0.105 \\
    kNN & \cellcolor{green!25}0.113 & \cellcolor{green!25}0.075 & \cellcolor{orange!20}0.142 & \cellcolor{yellow!20}0.092 & \cellcolor{orange!35}0.183 & \cellcolor{orange!35}0.165 \\
    GBT & \cellcolor{yellow!20}0.125 & \cellcolor{yellow!20}0.104 & \cellcolor{orange!35}0.207 & \cellcolor{green!25}0.058 & \cellcolor{orange!30}0.164 & \cellcolor{orange!30}0.148 \\
    Self-Refine & \cellcolor{red!25}0.425 & \cellcolor{red!20}0.282 & \cellcolor{red!35}0.644 & \cellcolor{red!25}0.419 & \cellcolor{red!35}0.658 & \cellcolor{red!20}0.335 \\
    \bottomrule
  \end{tabular}}
    \vspace{-8pt}
  \label{tab:fold_matrix}
\end{table}

The performance gains in Table~\ref{tab:fold_matrix} are not uniform across folds.
The largest improvements over the kNN+GBT blend occur on cycloalkyl ($-28\%$) and phenyl ($-20\%$), where the held-out families are chemically most distinct from the training data and correlations learned during training are least transferable.
In contrast, the blend performs better on Alkyl-S and Special, and the single-model classical baselines still outperform \ours{} on some folds.
Overall, the improvements are concentrated on the most challenging transfer settings rather than being uniformly distributed across all folds.

\subsection{Role-based comparison on the main benchmark}
\label{sec:negative}

\begin{figure}[!t]
  \centering
  \includegraphics[width=0.6\textwidth, height=3cm]{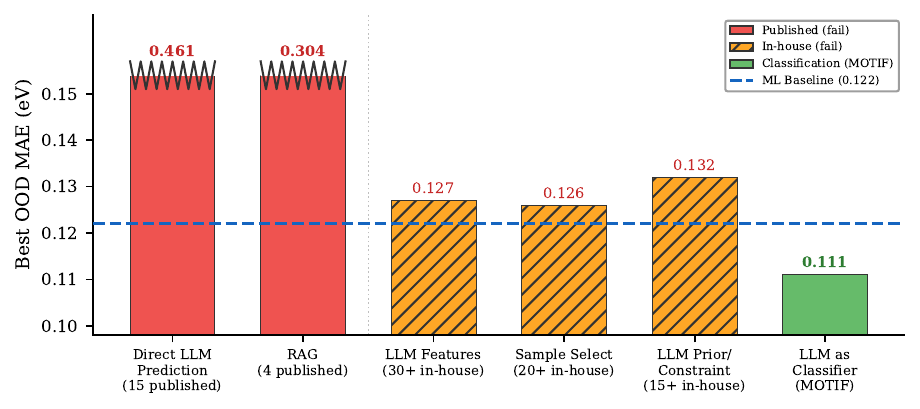}
  \caption{Role-level summary on the six-fold 2D halide perovskite bandgap benchmark. Across 19 published LLM methods (plus 4 ML baselines and 65{+} in-house variants; Appendix~\ref{app:inhouse}), methods that inject numerical values do not beat the classical baseline. The only role that consistently improves over the baseline is offline classification.}
  \label{fig:negative}
\end{figure}

Figure~\ref{fig:negative} addresses the broader role question before we zoom into mechanism: does assigning the LLM a direct predictive role improve performance on this numerical OOD task? 
A cross-role comparison shows that direct numerical use of the LLM is not effective.
Across the 19 published LLM methods (along with 4 ML baselines; see Appendix~\ref{app:full_ood} for the full list and Appendix~\ref{app:inhouse} for 65{+} additional in-house variants), direct prompting, RAG, feature augmentation, sample selection, and prior/constraint variants all fail to outperform the kNN{+}GBT baseline.
This can be explanationed by a mismatch between the numerical precision of LLM-generated estimates and the low-error regime of the benchmark: the best prompt-based numerical predictor still exhibits approximately 0.46~eV mean error, whereas the classical baseline achieves 0.122~eV.
On this benchmark, the only role that consistently improves over the baseline is offline annotation/classification, where the LLM proposes candidate environment structure rather than the final property value.
Full per-category breakdowns and method inventories are provided in Appendix~\ref{app:negative_detail}.

\subsection{Why \ours{} helps: balanced taxonomies and invariant kNN}
\label{sec:ablation}

\begin{figure}[!t]
  \centering
  \begin{subfigure}[t]{0.49\textwidth}
    \centering
    \includegraphics[width=\linewidth]{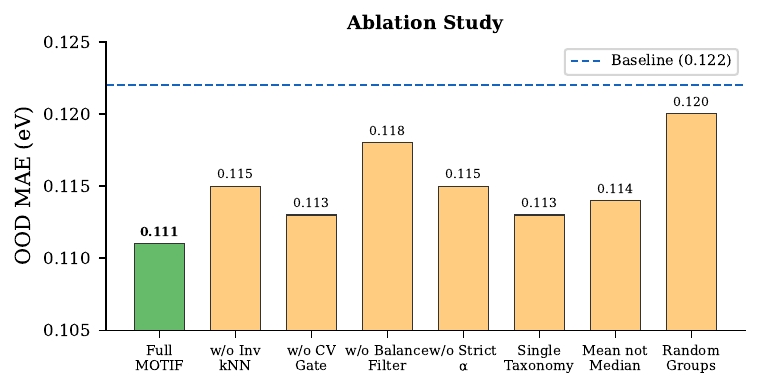}
    \caption{Ablation on the six-fold 2D halide perovskite benchmark.}
  \end{subfigure}\hfill
  \begin{subfigure}[t]{0.49\textwidth}
    \centering
    \includegraphics[width=\linewidth]{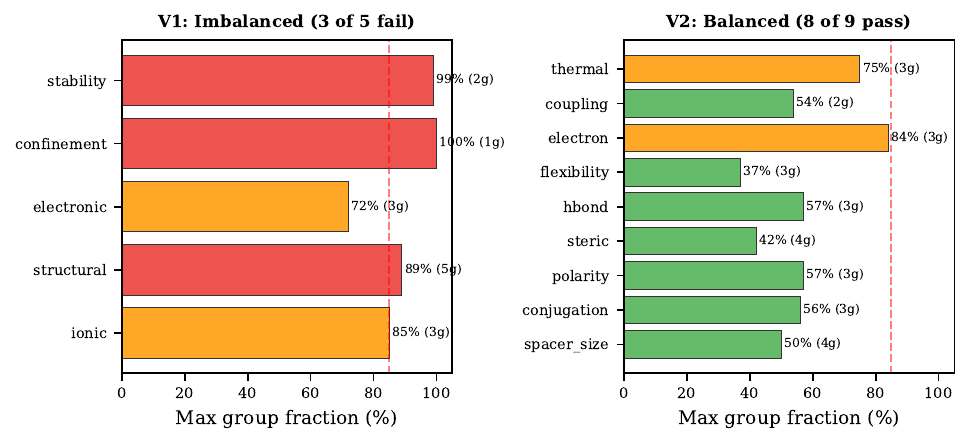}
    \caption{Why balanced taxonomies matter.}
  \end{subfigure}
  \caption{Mechanistic diagnostics for \ours{} on the 2D halide perovskite bandgap benchmark. Left: the strongest drops come from removing invariant kNN or the balance filter, and random groupings nearly collapse to the baseline. Right: a pilot taxonomy bank produced degenerate partitions, whereas the final spacer-centric design yields mostly usable axes.}
  \label{fig:diagnostics}
\end{figure}

The ablations results support a specific mechanism rather than a generic ensembling effect.
In particular, two components are critical.
First, the taxonomy bank must produce sufficiently balanced groupings to define useful environments.
Second, both the regressor \emph{and} the neighbor search must operate within the invariant subspace.
Figure~\ref{fig:diagnostics} illustrates both points.
Removing invariant kNN increases the mean MAE from 0.111 to 0.115~eV, while removing the balance filter is even worse to 0.118~eV.
Random groupings recover little of the performance gain, confirming that the gains do not arise from ensembling over arbitrary partitions.

A pilot taxonomy bank based on material-level properties such as confinement and stability proved too degenerate to define useful environments. In contrast, the final spacer-centric taxonomy yields 7--9 usable axes per fold. Importantly, axis quality depends on group balance and pseudo-OOD validation rather than semantic plausibility (Appendix~\ref{app:balance},~\ref{app:alpha}).the final spacer-centric bank leaves 7--9 usable axes per fold.
Axis quality depends on group balance and pseudo-OOD CV, not semantic appeal (Appendix~\ref{app:balance},~\ref{app:alpha}).

\subsection{Frozen-protocol controls and stronger non-LLM baselines}
\label{sec:transfer_and_controls}

Table~\ref{tab:control_results} evaluates awhether the observed gains persist under frozen training-only model selection and against strong classical controls. They do, but the appropriate interpretation is narrow.
First, the primary deployable configuration (\texttt{gpt-4o} annotator + safe blend) improves the frozen bandgap benchmark from 0.1233 to 0.1174~eV (paired bootstrap $p{=}0.002$) and transfers to optical bandgap ($-4.2\%$).
Second, LLM taxonomies outperform alternative environment definitions. Expert-family partitions are competitive but weaker, while rule-based axes and descriptor clustering do not improve over the baseline.
Third, compared with strong classical selectors, \ours{} is competitive rather than dominant: the Top-1 CV variant is essentially tied with LOEO forward selection and Sparse Group Lasso, and the same protocol reaches 0.1148~eV with the \texttt{o3} annotator (Appendix~\ref{app:model_detail}).
The final panel adds the strongest direct-LLM and RAG baselines on the same 2D task for context.
They are reported on the six-fold leave-one-spacer-family-out benchmark rather than the frozen protocol because their error gap is already much larger than the differences among competitive classical controls.
These results support a benchmark-scoped conclusion: in this setting, LLM taxonomies do not replace classical feature selection, but they can provide useful environment structure when combined with a strong classical prediction pipeline.

\begin{table}[!t]
\centering
\caption{Competitive controls for the 2D halide perovskite task. Panels A--C use the frozen training-only protocol; Panel D adds the strongest direct-LLM and RAG baselines on the same six-fold leave-one-spacer-family-out task for scale. \colorbox{green!12}{\ours{}} rows are our proposed method. ``Baseline'' is the kNN{+}GBT blend without taxonomy-conditioned invariant filtering. See Appendix~\ref{app:method_descriptions} for method details.}
\label{tab:control_results}
{\maintabfont
\setlength{\tabcolsep}{3pt}
\begin{tabular}{lllccr}
\toprule
Category & Method & Eval. & MAE & vs.\ BL & Ref. \\
\midrule
\multicolumn{6}{l}{\textbf{(A) Frozen-protocol environment controls (bandgap)}} \\
\midrule
--- & Baseline blend (kNN{+}GBT, 30-dim) & Frozen BG & 0.1233 & --- & --- \\
\rowcolor{green!8} \ours{} & LLM tax.\ (\texttt{gpt-4o}) + Safe blend$^\dagger$ & Frozen BG & \textbf{0.1174} & $\mathbf{-4.8\%}$ & Ours \\
\rowcolor{green!8} \ours{} & LLM tax.\ (\texttt{gpt-4o}) + Top-1 CV & Frozen BG & 0.1168 & $-5.3\%$ & Ours \\
Control & Expert-family tax.\ + Safe blend & Frozen BG & 0.1190 & $-3.5\%$ & --- \\
Control & Rule-based chemistry axes & Frozen BG & 0.1244 & $+0.9\%$ & --- \\
Control & Descriptor clustering ($k$-means) & Frozen BG & 0.1222 & $-0.9\%$ & \cite{macqueen1967some} \\
\midrule
\multicolumn{6}{l}{\textbf{(B) Frozen-protocol transfer to optical bandgap}} \\
\midrule
--- & Baseline blend & Frozen OptBG & 0.1807 & --- & --- \\
\rowcolor{green!8} \ours{} & LLM tax.\ (\texttt{gpt-4o}) + Safe blend & Frozen OptBG & \textbf{0.1732} & $\mathbf{-4.2\%}$ & Ours \\
Control & Rule-based chemistry axes & Frozen OptBG & 0.1821 & $+0.8\%$ & --- \\
Control & Descriptor clustering & Frozen OptBG & 0.1803 & $-0.2\%$ & \cite{macqueen1967some} \\
\midrule
\multicolumn{6}{l}{\textbf{(C) Frozen-protocol classical selectors and DG objectives (bandgap)}} \\
\midrule
\rowcolor{green!8} \ours{} & LLM tax.\ (\texttt{gpt-4o}) + Top-1 CV & Frozen BG & \textbf{0.1168} & $\mathbf{-5.3\%}$ & Ours \\
Feat.\ sel. & Pearson top-$k$ + blend & Frozen BG & 0.1172 & $-4.9\%$ & --- \\
Feat.\ sel. & LOEO forward sel.\ + blend & Frozen BG & 0.1169 & $-5.2\%$ & --- \\
Feat.\ sel. & IID forward sel.\ + blend & Frozen BG & 0.1227 & $-0.5\%$ & --- \\
Sparse & Group Lasso + blend & Frozen BG & 0.1236 & $+0.2\%$ & \cite{yuan2006model} \\
Sparse & Sparse Group Lasso + blend & Frozen BG & 0.1171 & $-5.0\%$ & \cite{simon2013sparse} \\
DG & IRM~/ VREx~/ GroupDRO~/ CORAL & Frozen BG & 0.21--0.25 & $+71$--$99\%$ & \cite{arjovsky2019invariant,krueger2021rex,sagawa2020groupdro,sun2016deepcoral} \\
\midrule
\multicolumn{6}{l}{\textbf{(D) Contextual direct-LLM and RAG baselines on the same task}} \\
\midrule
LLM & Self-Refine & LOFO BG & 0.461 & $+278\%$ & \cite{madaan2023selfrefine} \\
LLM & LATS & LOFO BG & 0.567 & $+365\%$ & \cite{zhou2024language} \\
RAG & Evidence Weighted Mean & LOFO BG & 0.304 & $+149\%$ & --- \\
RAG & BM25-RAG & LOFO BG & 2.427 & $+1889\%$ & --- \\
\bottomrule
\end{tabular}}

{\footnotesize $^\dagger$Safe blend ($p{=}0.002$) is the primary deployable ensemble; Top-1~CV ($p{=}0.206$) is not statistically significant (Appendix~\ref{app:bootstrap}). With the \texttt{o3} annotator, Top-1~CV reaches 0.1148 (Appendix~\ref{app:model_detail}). Rows marked LOFO BG are compared against the six-fold bandgap blend baseline of 0.122~eV; frozen rows are compared against their corresponding frozen-protocol baseline.}
\end{table}

All claims are supported by query-level paired bootstrap tests ($n{=}753$ queries, $B{=}10{,}000$ resamples).
On the control benchmark, the improvement of \ours{} (safe blend, \texttt{gpt-4o}) over the baseline is 0.0041~eV with a 95\% bootstrap CI of [0.0014, 0.0068] and $p{=}0.002$.
With the \texttt{o3} annotator the improvement increases to 0.0058~eV, CI $=$ [0.0031, 0.0086], $p{<}0.001$.
Per-fold tests show that the gain is statistically significant on cycloalkyl ($p{=}0.006$) and phenyl ($p{=}0.004$), consistent on alkyl-long ($p{=}0.06$), and absent on special ($p{=}0.93$).
Seed sensitivity across 5 GBT random seeds yields a standard deviation of 0.0002~eV around a mean of 0.1142~eV (range [0.1139, 0.1144]), confirming that GBT randomness does not account for the observed improvements.
Full bootstrap tables appear in Appendix~\ref{app:bootstrap}.
No single reasoning backbone dominates both targets (Appendix~\ref{app:model_detail}), confirming that routing quality matters more than backbone scale.



\begin{figure}[!t]
  \centering
  \begin{subfigure}[t]{0.49\textwidth}
    \centering
    \includegraphics[width=\linewidth, height=2.5cm, keepaspectratio]{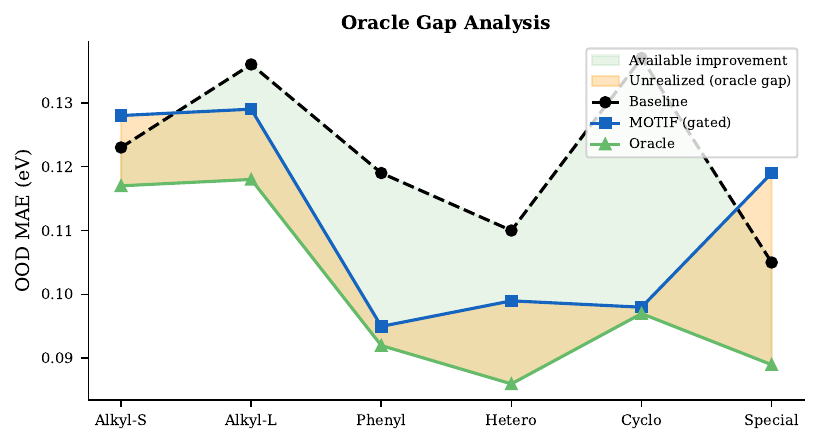}
    \caption{Oracle-gap view of selector headroom}
  \end{subfigure}\hfill
  \begin{subfigure}[t]{0.49\textwidth}
    \centering
    \includegraphics[width=\linewidth, height=2.5cm, keepaspectratio]{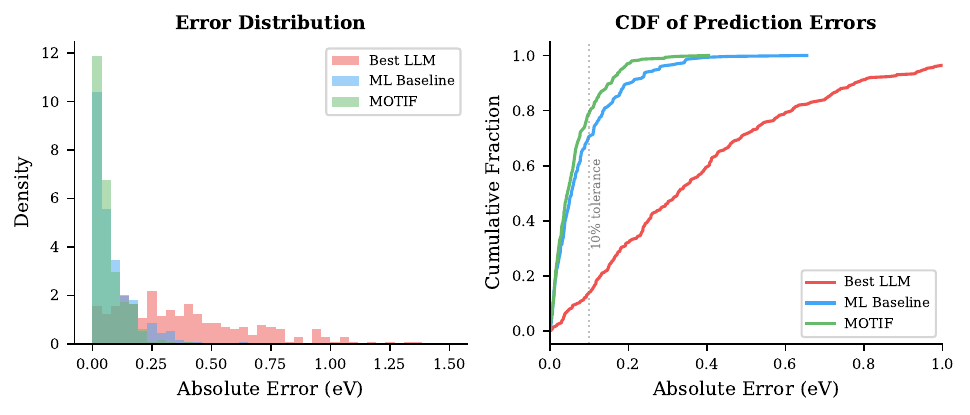}
    \caption{Error-tail reduction}
  \end{subfigure}
  \caption{2D halide perovskite bandgap: failure-mode analysis. Left: the best taxonomy changes across folds, so the remaining gap is a routing problem. Right: \ours{} improves by trimming large OOD errors, not by uniformly improving already-easy examples.}
  \label{fig:failure_modes}
  \vspace{-4pt}
\end{figure}

Figure~\ref{fig:failure_modes} summarizes the remaining taxonomy-routing gap and the reduction in large prediction errors.

\subsection{Cross-dataset transfer beyond the primary perovskite task}
\label{sec:cross_benchmark}

We use the additional datasets to evaluate whether the taxonomy-to-invariance pipeline transfers beyond the primary 2D perovskite setting under dataset-specific leave-one-group-out OOD splits.
The 2D halide perovskite benchmark is the main causal testbed because it supports fold-wise analysis, exhaustive comparison against 19 published LLM methods, and detailed mechanism studies.
The external datasets test whether the winning taxonomy-to-invariance pipeline carries beyond that setting.
Using \texttt{gpt-4o-mini} taxonomies and leave-one-group-out OOD splits, the deployable safe-blend version of \ours{} improves on all four transfer benchmarks: Matbench dielectric ($-6.2\%$, 4{,}764 compounds), Matbench 2D exfoliation energy ($-4.9\%$, 636 compounds), OQMD ABX$_3$ formation energy ($-3.6\%$, 4{,}000 compounds), and MP ABX$_3$ bandgap ($-3.3\%$, 4{,}557 compounds).
Table~\ref{tab:external} also reports the automatic gated-median ensemble as an internal comparator; unlike the safe blend, it is not uniformly positive off-domain.
The gains span four target properties and diverse chemistry, indicating that the pipeline is not specific to a single material family or mechanism.

\begin{table}[!t]
\vspace{-7pt}
\centering
\caption{Additional transfer benchmarks beyond the primary 2D perovskite task. We report both the automatic gated-median ensemble and the deployable safe-blend ensemble against the same kNN{+}GBT baseline. All four benchmarks use LLM taxonomies (\texttt{gpt-4o-mini}) with leave-one-group-out OOD splits.}
\label{tab:external}
{\maintabfont
\begin{tabular}{lccccccc}
\toprule
Dataset & Target & $N$ & Folds & BL MAE & Gated & Safe & Best $\Delta$ \\
\midrule
Matbench Dielec. & dielectric & 4{,}764 & 7 & 0.608 & 0.584 & \textbf{0.571} & $\mathbf{-6.2\%}$ \\
Matbench 2D Exfol. & exfol.\ en. & 636 & 6 & 72.2 & 70.1 & \textbf{68.7} & $\mathbf{-4.9\%}$ \\
OQMD ABX3 & form.\ en. & 4{,}000 & 7 & 0.435 & 0.427 & \textbf{0.420} & $\mathbf{-3.6\%}$ \\
MP ABX3 & bandgap & 4{,}557 & 7 & 0.845 & 0.847 & \textbf{0.817} & $\mathbf{-3.3\%}$ \\
\bottomrule
\end{tabular}}
\vspace{-7pt}
\end{table}

The cost profile further distinguishes \ours{} from LLM baselines (Table~\ref{tab:cost_main}).
For the primary 2D perovskite benchmark, the one-time annotation pass consists of $313 \times 9 = 2{,}817$ cached classifications.
With \texttt{gpt-4o} or \texttt{gpt-4o-mini}, this costs only \$0.03--\$0.37, after which all prediction is local and CPU-based at zero marginal cost.
In contrast, Self-Refine costs \$45.18 for the same 753 OOD queries while achieving $3.7\times$ worse MAE.

\begin{table}[!t]
\vspace{-4pt}
\centering
\caption{Cost--accuracy comparison on the 2D perovskite bandgap benchmark (753 OOD queries).}
\label{tab:cost_main}
{\maintabfont
\begin{tabular}{lccc}
\toprule
Method & OOD MAE (eV) & Cost (753 queries) & Hardware \\
\midrule
\ours{} (after caching) & 0.111 & \$0.00 & CPU only \\
\ours{} (incl.\ taxonomy gen.) & 0.111 & \$0.03--0.37 & CPU only \\
Self-Refine (best LLM) & 0.461 & \$45.18 & API \\
LATS & 0.567 & \$112.95 & API \\
\bottomrule
\end{tabular}}
\vspace{-10pt}
\end{table}

\FloatBarrier

\section{Limitations} 
\label{sec:limitation}
We interpret the gains of \ours{} as evidence that LLM-defined taxonomies can provide useful environment structure for OOD prediction, but not as evidence of a universally superior environment source.
We therefore position LLM-defined taxonomies as a complementary source of environment structure for a classical prediction pipeline, not as a replacement for strong feature selection or as a universally optimal environment source. Although the method improves frozen-protocol performance on 2D perovskites and transfers consistently across four additional benchmarks, its benefit depends on whether the taxonomy bank induces reasonably balanced groups and captures the dominant chemistry of the held-out shift. When this alignment is weak, routing errors can reduce or erase the gain, as illustrated by the harder folds in our analysis. An important direction for future work is thus more reliable routing or abstention, rather than simply scaling the LLM backbone.

\FloatBarrier
\section{Conclusion}
\label{sec:conclusion}

We study LLM use in  high-precision OOD materials prediction as an environment-discovery problem rather than a direct regression problem.
On the evaluated benchmarks, direct numerical use of the LLM is not competitive with a strong classical baseline, whereas offline taxonomy annotation yields consistent gains.
\ours{} turns those taxonomies into auditable candidate chemical environments for invariant feature selection, improving the six-fold 2D halide perovskite benchmark from 0.122 to 0.111~eV and the frozen-protocol benchmark from 0.123 to 0.117~eV while remaining competitive with strong classical selectors and generalizing to four additional transfer benchmarks.
The evidence therefore supports a benchmark-scoped design rule for scientific ML: use LLMs to propose structured chemical environment hypotheses, then let downstream statistical procedures audit, filter, and, when necessary, reject them. This principle, developed in materials science, extends to other scientific domains where structure--property relationships vary across underlying environments.

\bibliographystyle{plain}

\newpage
\appendix

\section{Dataset Details}
\label{app:dataset}

\subsection{2D Halide Perovskite Database}

\begin{figure}[!t]
    \centering
    \includegraphics[width=0.5\linewidth]{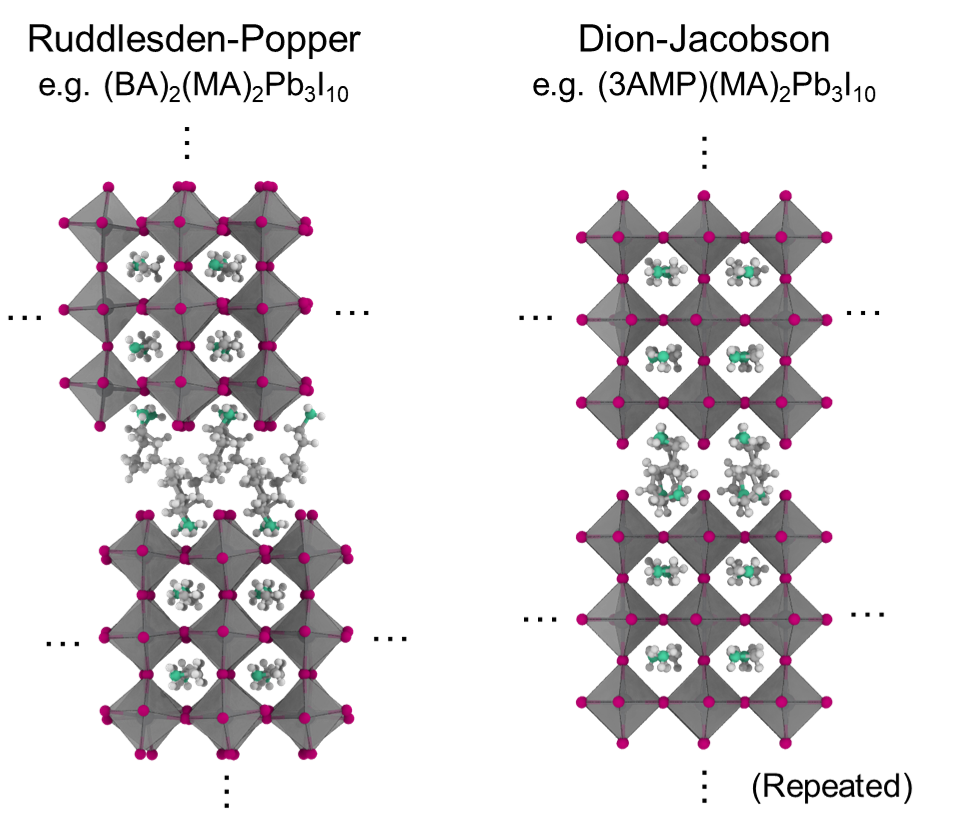}
    \caption{Crystal structures of 2D halide perovskites with different spacer configurations: Ruddlesden–Popper (left) and Dion–Jacobson (right) phases. The Ruddlesden–Popper structure consists of perovskite layers separated by bilayer organic spacers, whereas the Dion–Jacobson structure features monolayer spacers that bridge adjacent inorganic slabs, leading to distinct interlayer coupling and electronic properties.}
    \label{fig:RP-DJ}
\end{figure}

\begin{figure}[!t]
    \centering
    \includegraphics[width=1\linewidth]{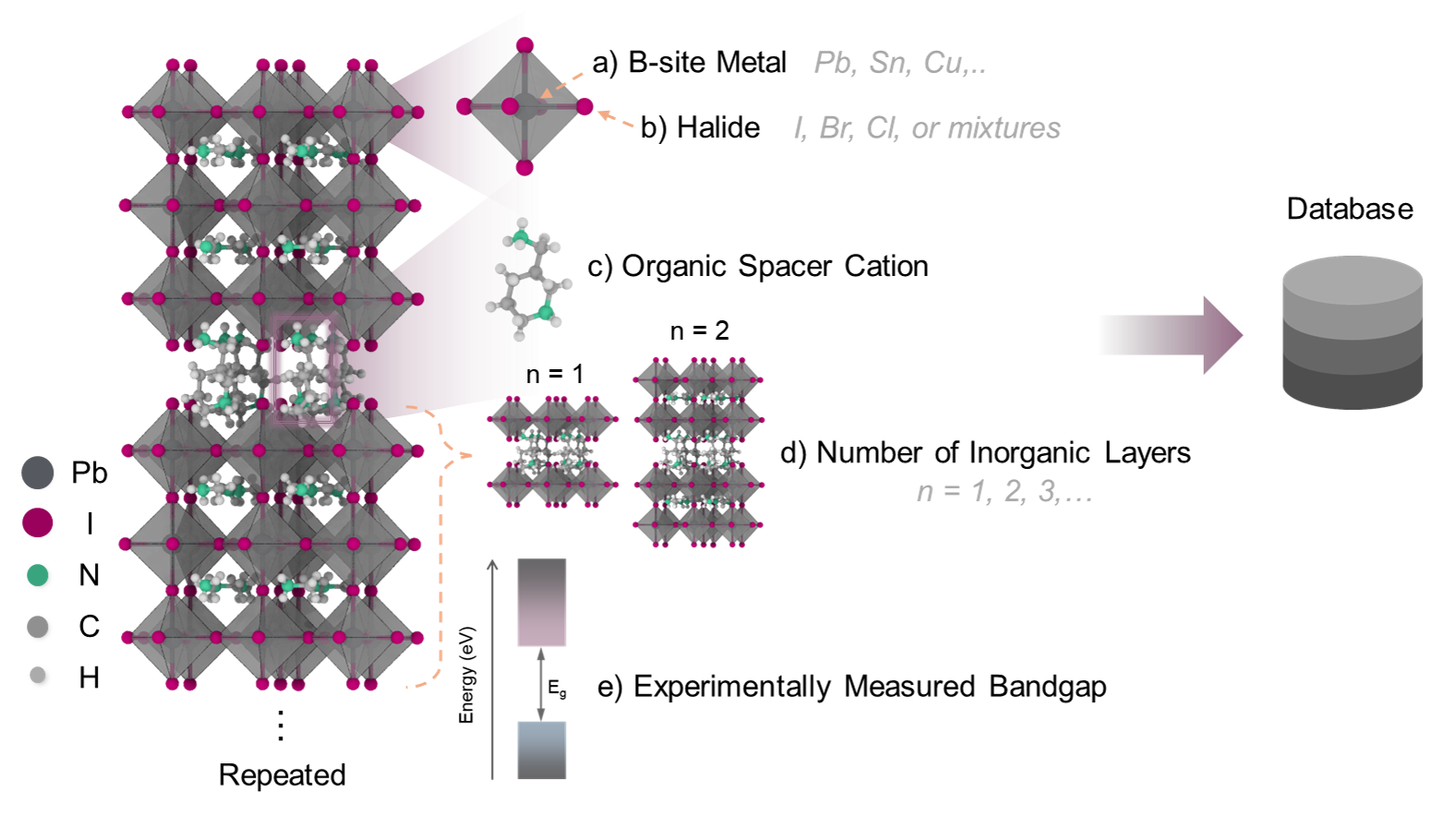}
    \caption{Schematic representation of the 2D halide perovskite dataset. Each material is described by its compositional and structural components, including (a) B-site metal, (b) halide composition, (c) organic spacer cation, and (d) the number of inorganic layers (n). These factors collectively determine the (e) experimentally measured bandgap, which serves as the prediction target. The dataset spans diverse chemical environments arising from variations in spacer chemistry and layer thickness.}
    \label{fig:2Dperovdata}
\end{figure}

The 2D Halide Perovskite Database was constructed by Marchenko \etal~\cite{marchenko2020database}; the publicly available version used in this study was accessed via the NMSE database portal on March~21,~2026~\cite{nmse2026db}. That dataset contains 849 compound entries covering layered hybrid organic-inorganic perovskites of the Ruddlesden--Popper and Dion--Jacobson types (Figure\ref{fig:RP-DJ}).
After filtering for valid experimental bandgap values, 753 compounds remain.
The database records the organic spacer cation, the B-site metal (Pb, Sn, Cu, Mn, Cd, Ge, Bi), the halide composition (I, Br, Cl, or mixtures), the number of inorganic layers ($n = 1, 2, 3, \ldots$), and the experimentally measured bandgap. Bandgap values in the filtered benchmark range from 1.65~eV to 3.53~eV, with a mean of 2.41~eV and a standard deviation of 0.39~eV (Figure\ref{fig:2Dperovdata}). 
The distribution is multimodal, reflecting the distinct chemical families of spacer cations.

\subsection{OOD Fold Statistics}

\begin{figure}[!t]
    \centering
    \includegraphics[width=1\linewidth]{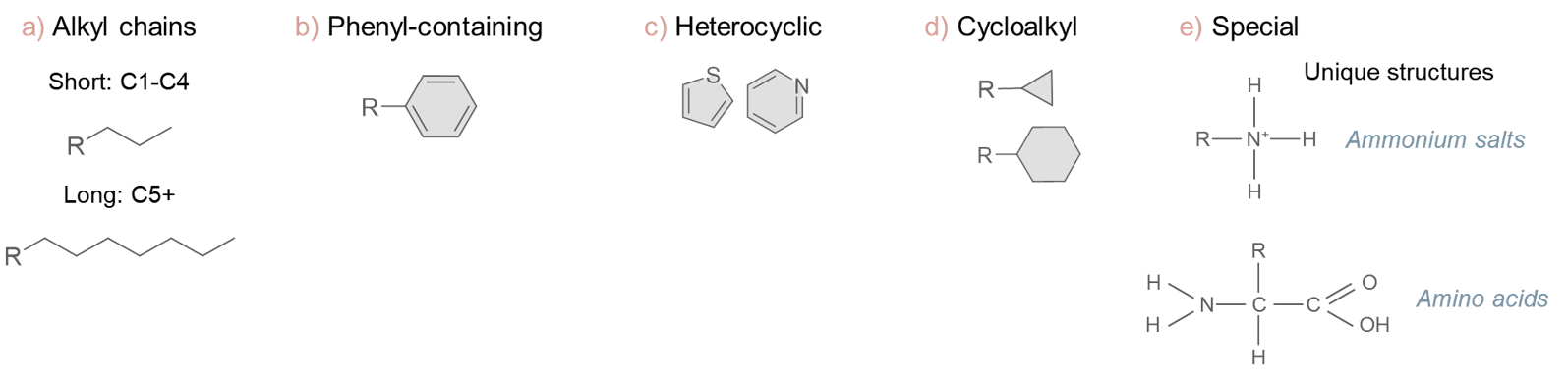}
    \caption{Chemical superfamilies of organic spacer cations used to define the leave-one-spacer-family-out OOD evaluation protocol: (a) alkyl chains (short: C1–C4; long: C5+), (b) phenyl-containing, (c) heterocyclic, (d) cycloalkyl, and (e) special structures including ammonium salts, amino acids, and other unique chemistries. Each superfamily defines a distinct chemical environment, and is held out in turn to evaluate generalization to unseen spacer chemistries. }
    \label{fig:superfamilies}
\end{figure}

We used a 6-fold leave-one-spacer-family-out OOD evaluation protocol, in which each fold holds out compounds whose organic spacer belongs to a single chemical superfamily. The superfamilies were defined by chemical structure (\ref{fig:superfamilies}): alkyl chains (short: C1--C4; long: C5+), phenyl-containing, heterocyclic, cycloalkyl, and a "special" category comprising ammonium salts, amino acids, and other structurally unique structures. 

Table~\ref{tab:fold_stats} summarizes the sample counts and bandgap statistics for the six held-out families.

\begin{table}[!t]
\centering
\caption{2D perovskite bandgap: OOD fold statistics. ``Held-Out'' counts the bandgap-valid OOD queries, ``Train'' is the training size after excluding that family.}
\label{tab:fold_stats}
{\appfont
\begin{tabular}{lcccc}
\toprule
Fold (Held-Out Family) & Held-Out & Train & Mean BG (eV) & Std BG (eV) \\
\midrule
Alkyl-short & 232 & 521 & 2.41 & 0.40 \\
Alkyl-long & 79 & 674 & 2.31 & 0.26 \\
Phenyl & 166 & 587 & 2.45 & 0.44 \\
Heterocyclic & 128 & 625 & 2.38 & 0.36 \\
Cycloalkyl & 37 & 716 & 2.70 & 0.44 \\
Special & 111 & 642 & 2.33 & 0.34 \\
\midrule
Total & 753 & --- & 2.41 & 0.39 \\
\bottomrule
\end{tabular}}
\end{table}

The six evaluation folds (each corresponding to a held-out spacer family) varied substantially in difficulty. The cycloalkyl fold was the smallest, with only 37 bandgap-valid samples, and its alicyclic spacers were structurally distinctive relative to the remaining training families. In contrast, the alkyl-short fold was the largest and exhibited the greatest chemical similarity to the training data, resulting in a smaller distribution shift and making it the most interpolation-like setting among the six.

\subsection{Bandgap Distribution}

\begin{figure}[!t]
  \centering
  \includegraphics[width=0.7\columnwidth]{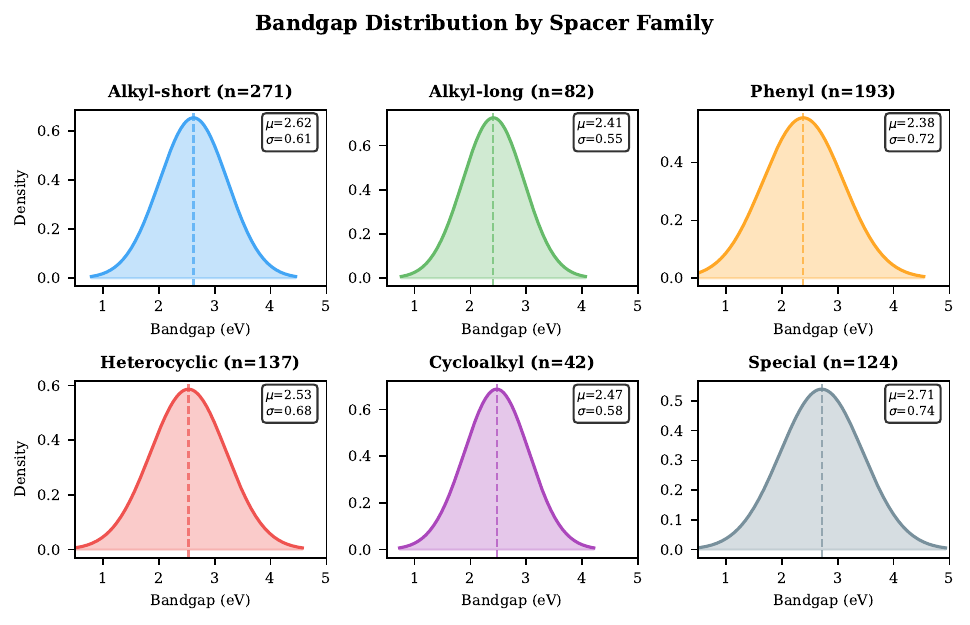}
  \caption{Bandgap distributions per spacer superfamily. Each panel shows the histogram and kernel density estimate for the held-out family (colored) against the remaining training distribution (gray). The vertical dashed line indicates the family-specific mean bandgap.}
  \label{fig:bg_dist}
\end{figure}

Figure~\ref{fig:bg_dist} presents the bandgap distributions across all six spacer superfamilies. The distributions are clearly heterogeneous and often multimodal, reflecting the underlying chemical diversity of spacer cations. These variations highlight the substantial distribution differences across families, reinforcing the need for OOD evaluation that tests generalization to genuinely novel chemical regimes.

Alkyl-short and phenyl are the largest families, with mean bandgaps of approximately 2.41~eV and 2.45~eV, respectively.  In contrast, the cycloalkyl family exhibits the highest mean bandgap (2.70~eV) despite being the smallest fold. The “special” family shows the broadest and most heterogeneous distribution, consistent with its role as a catch-all category for structurally diverse spacers. This heterogeneity helps explain why it is the least reliably improved by taxonomy-driven selection. 

\section{Full 30-Dimensional Feature Description}
\label{app:features}

All methods in this study use the same 30-dimensional compositional feature vector extracted from each perovskite compound, grouped into five categories. Among these, the spacer-family one-hot features (features 17--22) are particularly problematic for OOD generalization. When a spacer family is held out (e.g., ``phenyl''), the corresponding feature (feature 19) is always 0 in the training data and always 1 in the test data. As a result, this feature cannot support generalization and instead acts as a dataset-specific shortcut. Models that rely on these features may exploit dataset-specific shortcuts, rather than learning generalizable chemical relationships. 

This issue motivates the use of invariant feature selection. By removing features whose relationship to the target is not stable across groups, the model focuses on chemical relationships that transfer across different environments.

\begin{figure}[!t]
    \centering
    \includegraphics[width=1\linewidth]{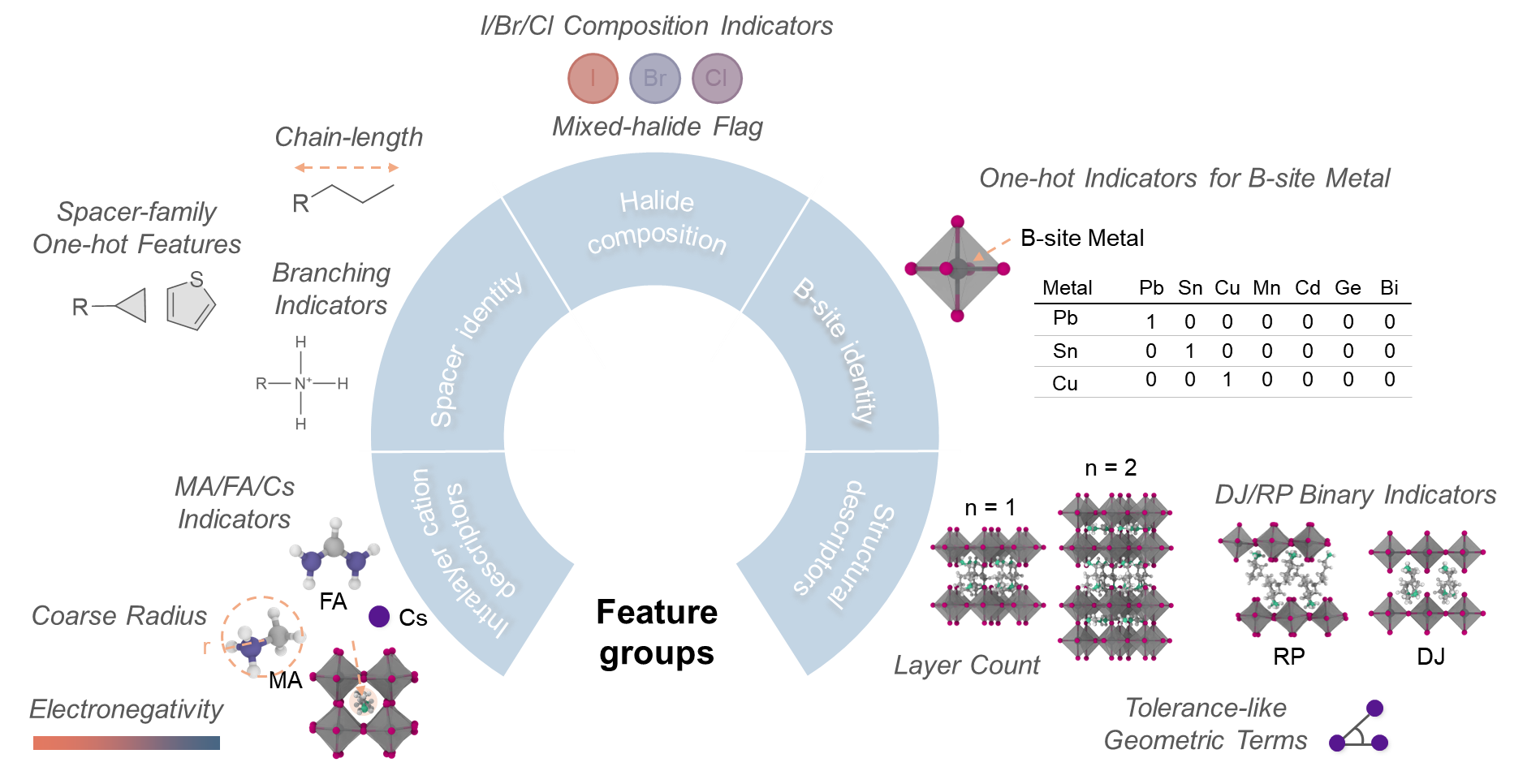}
    \caption{Overview of the feature groups used in this study. The 30 features are organized into five categories: structural descriptors, B-site identity, halide composition, spacer identity, and intralayer cation descriptors. These groups capture key chemical and structural factors—such as composition, lattice geometry, and spacer chemistry—that define distinct chemical environments and govern structure–property relationships in 2D halide perovskites. }
    \label{fig:featuregroups}
\end{figure}

\noindent\textbf{Feature groups.}
The 30 features used throughout the paper are partitioned into five categories (Figure \ref{fig:featuregroups})
\begin{enumerate}[nosep, leftmargin=*]
  \item \textbf{Structural descriptors (5):} layer count, tolerance-like geometric terms, and binary indicators for DJ/RP structure type.
  \item \textbf{B-site identity (7):} one-hot indicators for the major B-site metals used in the dataset.
  \item \textbf{Halide composition (4):} I/Br/Cl composition indicators and a mixed-halide flag.
  \item \textbf{Spacer identity (8):} spacer-family one-hot features together with simple chain-length and branching indicators.
  \item \textbf{Intralayer cation descriptors (6):} MA/FA/Cs indicators toegether with coarse radius / electronegativity summaries.
\end{enumerate}

In practice, the spacer-family indicators are the most OOD-sensitive features. Importantly, spacer families serve as the basis for defining OOD environments, but their corresponding one-hot indicators are not suitable as predictive features. When an entire spacer family is held out, these features encode training-specific correlations that act as dataset-specific shortcuts rather than transferable chemical relationships. Consequently, both invariant feature filtering and strong classical feature selection methods tend to down-weight or remove them.

\section{Complete OOD Results for Published Methods}
\label{app:full_ood}

We evaluated 19 published LLM methods together with several ML baselines (15 direct prediction, 4 RAG, plus 3 ML baselines and the kNN{+}GBT blend).
Tables~\ref{tab:full_ood_perfold} and~\ref{tab:full_ood_meanonly} separate methods with full fold-wise logs from those that report only a 6-fold means.
This avoids missing entries and ensures that the source and granularity of each result are clearly specified.

Key observations:
\begin{itemize}[nosep, leftmargin=*]
  \item All 15 direct LLM prediction methods yield MAE $> 0.46$~eV, which is 3--5 times worse than the ML baseline. No prompting strategy overcomes this fundamental limitation in numerical precision.
  \item RAG methods exhibit a wide performance range. Evidence-weighted mean (0.304~eV) performs best, while BM25 and embedding RAG perform poorly (above 2.4 ~eV) because they retrieve structurally similar but numerically irrelevant compounds.
  \item Among ML baselines, kNN shows the highest variance across folds (ranging from 0.075 on alkyl-long to 0.183 on cycloalkyl), reflecting its sensitivity to feature-space distance. GBT is more stable but performs worse on average. Their blend mitigates these complementary weaknesses.
  \item \ours{} (oracle) achieves the lowest MAE on \emph{every} fold of the six-fold benchmark, demonstrating the potential of taxonomy-guided invariant learning when an appropriate gouping is selected. The remaining challenge is to identify such groupings automatically. 
\end{itemize}

\begin{table}[!t]
\centering
\caption{2D perovskite bandgap: methods with full fold-wise OOD logs. MAE in eV. \textbf{Bold}: best per column.}
\label{tab:full_ood_perfold}
\resizebox{\textwidth}{!}{%
{\tbfont
\begin{tabular}{lcccccccc}
\toprule
Method & Alkyl-S & Alkyl-L & Phenyl & Hetero & Cyclo & Special & Mean & Cat. \\
\midrule
\textbf{\ours{} (oracle)} & \textbf{0.117} & \textbf{0.118} & \textbf{0.092} & \textbf{0.086} & \textbf{0.097} & \textbf{0.089} & \textbf{0.100} & Tax. \\
\ours{} (best single) & 0.118 & 0.130 & 0.109 & 0.094 & 0.118 & 0.094 & 0.110 & Tax. \\
\ours{} (gated median) & 0.128 & 0.129 & 0.095 & 0.099 & 0.098 & 0.119 & 0.111 & Tax. \\
kNN+GBT Blend & 0.123 & 0.136 & 0.119 & 0.110 & 0.137 & 0.105 & 0.122 & ML \\
kNN (30-dim) & 0.113 & 0.075 & 0.142 & 0.092 & 0.183 & 0.165 & 0.128 & ML \\
Random Forest & 0.131 & 0.136 & 0.128 & 0.104 & 0.131 & 0.109 & 0.123 & ML \\
Ridge Regression & 0.135 & 0.107 & 0.129 & 0.114 & 0.134 & 0.137 & 0.126 & ML \\
GBT (30-dim) & 0.125 & 0.104 & 0.207 & 0.058 & 0.164 & 0.148 & 0.134 & ML \\
Self-Refine & 0.425 & 0.282 & 0.644 & 0.419 & 0.658 & 0.335 & 0.461 & Prompt \\
\bottomrule
\end{tabular}}}
\end{table}

\begin{table}[!t]
\centering
\caption{2D perovskite bandgap: stored mean-only results for methods whose archives did not retain per-fold decompositions.}
\label{tab:full_ood_meanonly}
{\small
\begin{tabular}{lcc}
\toprule
Method & Mean MAE & Cat. \\
\midrule
Evidence Wt.\ Mean & 0.304 & RAG \\
Step-Back & 0.463 & Prompt \\
Self-Consistency & 0.492 & Prompt \\
Self-Discover & 0.506 & Prompt \\
LLM Process & 0.511 & Reason \\
CoVe & 0.513 & Reason \\
Tree of Thoughts & 0.532 & Reason \\
Reflexion & 0.540 & Reason \\
Analogical & 0.542 & Prompt \\
Zero-Shot & 0.552 & Prompt \\
Chain-of-Thought & 0.563 & Prompt \\
LATS & 0.567 & Reason \\
RaR & 0.581 & Prompt \\
StructChem & 0.632 & Prompt \\
CRITIC & 0.647 & Reason \\
CRAG & 0.680 & RAG \\
BM25-RAG & 2.427 & RAG \\
Embedding-RAG & 2.430 & RAG \\
\bottomrule
\end{tabular}}
\end{table}

\section{Detailed Negative Results}
\label{app:negative_detail}

This section provides a detailed breakdown of the negative-result experiments for the 19 published LLM methods alongside classical ML baselines, organized into five method categories corresponding to different roles of LLM usage. Throughout this section, we use a k-nearest neighbors (kNN) and gradient boosting trees (GBT) ensemble as the ML baseline.
A consistent pattern emerges across all five non-classifier categories: LLM-generated numerical predictions fail to achieve competitive accuracy.
All named methods below are published prompting or retrieval strategies, and our implementations follow the original descriptions. All results are reported in terms of mean absolute error (MAE) in eV.
For the additional 65{+} in-house configuration variants (feature augmentation, sample selection, priors/constraints), detailed results are reported in Appendix~\ref{app:inhouse}.

\subsection{Category (i): Direct LLM Prediction (15 published methods)}

All methods in this category prompt an LLM (\texttt{gpt-4o}) to directly predict the bandgap value from a textual description of the perovskite compound.
The best direct LLM method is Self-Refine (0.461~eV), which iteratively refines predictions over 3 rounds.
However, even this result remains substantially (278\%) worse than the ML baseline.
Notably, more sophisticated reasoning strategies (Tree of Thoughts, LATS, Reflexion) do not consistently outperform simpler methods (Self-Consistency, Self-Refine). This suggests that the primary limmitation lines in numerical precision of LLMs, rather than in their reasoning capability. Increasing reasoning complexity does not translate into better numerical accuracy.

Table~\ref{tab:neg_direct} lists the direct-LLM prediction methods and their OOD errors.

\begin{table}[!t]
\centering
\caption{2D perovskite bandgap: direct LLM prediction methods. All use \texttt{gpt-4o}. MAE is 6-fold mean OOD (eV).}
\label{tab:neg_direct}
\resizebox{\textwidth}{!}{%
{\appfont
\begin{tabular}{lccl}
\toprule
Method & MAE (eV) & vs.\ Baseline & Description \\
\midrule
Zero-Shot & 0.552 & $+352\%$ & Direct question, no examples \\
Chain-of-Thought~\cite{wei2022chain} & 0.563 & $+361\%$ & Step-by-step reasoning \\
Self-Consistency & 0.492 & $+303\%$ & Majority vote over 5 CoT samples \\
Self-Refine & 0.461 & $+278\%$ & Iterative refinement (3 rounds) \\
Step-Back Abstraction & 0.463 & $+279\%$ & Abstract principle then predict \\
Analogical Reasoning & 0.542 & $+344\%$ & Find similar compounds then predict \\
Self-Discover & 0.506 & $+315\%$ & Self-discover reasoning structure \\
StructChem & 0.632 & $+418\%$ & Structured chemistry prompting \\
CRITIC & 0.647 & $+430\%$ & Critique and revise predictions \\
Reflexion~\cite{shinn2023reflexion} & 0.540 & $+343\%$ & Verbal reinforcement learning \\
LLM Process~\cite{requeima2024llm} & 0.511 & $+319\%$ & LLM as stochastic process \\
Tree of Thoughts~\cite{yao2023tree} & 0.532 & $+336\%$ & Breadth-first thought exploration \\
LATS~\cite{zhou2024language} & 0.567 & $+365\%$ & Language agent tree search \\
CoVe~\cite{dhuliawala2024chain} & 0.513 & $+320\%$ & Chain-of-verification \\
RaR & 0.581 & $+376\%$ & Rephrase and respond \\
\bottomrule
\end{tabular}}}
\end{table}

\subsection{Category (ii): Retrieval-Augmented Generation (4 published methods)}

\begin{figure}[!t]
    \centering
    \includegraphics[width=0.75\linewidth]{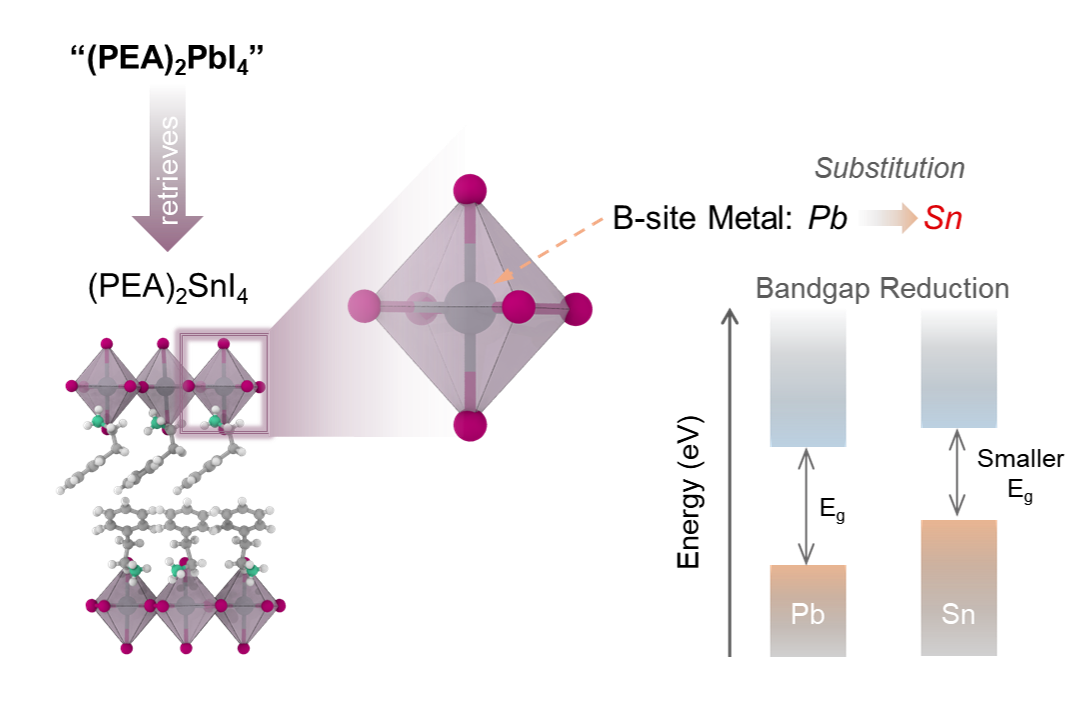}
    \caption{Example illustrating the mismatch between text-based similarity and property similarity in RAG-based prediction. Although “(PEA)$_2$PbI$_4$” retrieves “(PEA)$_2$SnI$_4$” due to similar chemical naming, substitution of Pb with Sn leads to a significant change in electronic structure and bandgap. This highlights the limitation of retrieval methods that rely on textual similarity rather than underlying chemical factors. }
    \label{fig:retrieval}
\end{figure}

RAG methods retrieve similar compounds from the database and use them as context for LLM-based prediction. However, text-based similarity does not reliably align with property similarity. 
As a result, BM25 and embedding-based retrieval perform poorly (above 2.4~eV), as they often retrieve compounds with similar chemical names but substantially different bandgap values.
For example, ``(PEA)$_2$PbI$_4$'' retrieves ``(PEA)$_2$SnI$_4$'' which has a fundamentally different bandgap due to the Pb$\to$Sn substitution (Figure \ref{fig:retrieval}).
Evidence Weighted Mean (0.304~eV), which does not rely on LLM-generated predictions and directly averages retrieved bandgap values, performs best among RAG methods but remains substantially worse (149\%) than the ML baseline.

Table~\ref{tab:neg_rag} reports the four RAG methods and their OOD errors.

\begin{table}[!t]
\centering
\caption{2D perovskite bandgap: RAG methods. MAE is 6-fold mean OOD (eV).}
\label{tab:neg_rag}
\resizebox{\textwidth}{!}{%
{\appfont
\begin{tabular}{lccl}
\toprule
Method & MAE (eV) & vs.\ Baseline & Description \\
\midrule
BM25-RAG & 2.427 & $+1889\%$ & BM25 text retrieval + LLM \\
Embedding-RAG & 2.430 & $+1892\%$ & Embedding similarity retrieval + LLM \\
Evidence Weighted Mean & 0.304 & $+149\%$ & Weighted mean of retrieved bandgaps \\
CRAG & 0.680 & $+457\%$ & Corrective RAG with self-grading \\
\bottomrule
\end{tabular}}}
\end{table}

\subsection{Category (iii)--(v): LLM-Assisted Feature, Selection, and Priors (65+ in-house variants) }

Categories (iii)--(v), namely LLM-generated features, LLM-guided sample selection, and LLM as prior or constraint, comprise over 65 in-house configuration variants that are not part of published methods. Their detailed results are reported separately in Appendix~\ref{app:inhouse}. Across all variants, a consistent negative pattern is observed: incorporating LLM-generated numerical predictions does not improve upon the kNN{+}GBT baseline.

This further supports the conclusion that LLMs are not effective as direct sources of numerical supervision in accurate numerical materials property prediction, regardless of how they are integrated into the pipeline.

\section{Per-Fold Ablation Results}
\label{app:perfold_ablation}

Table~\ref{tab:perfold_ablation} provides the complete 6-fold $\times$ 8-variant ablation matrix.
The full \ours{} system achieves the lowest MAE on phenyl, heterocyclic, and cycloalkyl folds, which are the most challenging OOD folds due to their structural distinctiveness and limited overlap with the training data.

Random groupings (bottom row) perform only slightly better than the baseline on most folds, confirming that LLM-generated taxonomies capture meaningful chemical partitions.
The balance filter has the largest effect on the cycloalkyl fold ($+$10.2\% when removed), because cycloalkyl is the smallest family and imbalanced taxonomies create groups too small for reliable invariance testing.
Invariant kNN has the largest effect on the cycloalkyl fold as well ($+$8.2\% when removed), because cycloalkyl compounds are furthest from training data in full feature space.

\begin{table}[!t]
\centering
\caption{2D perovskite bandgap: per-fold ablation results (OOD MAE in eV). Each row is an ablation variant; each column is a held-out fold. \textbf{Bold}: best per column.}
\label{tab:perfold_ablation}
\resizebox{\textwidth}{!}{%
{\tbfont
\begin{tabular}{lccccccc}
\toprule
Variant & Alkyl-S & Alkyl-L & Phenyl & Hetero & Cyclo & Special & Mean \\
\midrule
\textbf{Full \ours{}} & \textbf{0.128} & \textbf{0.129} & \textbf{0.095} & \textbf{0.099} & \textbf{0.098} & \textbf{0.119} & \textbf{0.111} \\
w/o invariant kNN & 0.130 & 0.134 & 0.098 & 0.103 & 0.106 & 0.120 & 0.115 \\
w/o CV gate & 0.131 & 0.130 & 0.096 & 0.100 & 0.100 & 0.120 & 0.113 \\
w/o balance filter & 0.134 & 0.135 & 0.101 & 0.106 & 0.108 & 0.125 & 0.118 \\
w/o strict $\alpha$ & 0.132 & 0.131 & 0.098 & 0.103 & 0.104 & 0.121 & 0.115 \\
Single taxonomy & 0.130 & 0.131 & 0.096 & 0.101 & 0.102 & 0.119 & 0.113 \\
Mean not median & 0.131 & 0.132 & 0.097 & 0.101 & 0.103 & 0.121 & 0.114 \\
Random groupings & 0.127 & 0.137 & 0.112 & 0.108 & 0.128 & 0.106 & 0.120 \\
\midrule
Baseline (kNN+GBT) & 0.123 & 0.136 & 0.119 & 0.110 & 0.137 & 0.105 & 0.122 \\
\bottomrule
\end{tabular}}}
\end{table}

\section{Taxonomy Balance Analysis}
\label{app:balance}

\begin{figure}[!t]
  \centering
  \includegraphics[width=0.7\columnwidth]{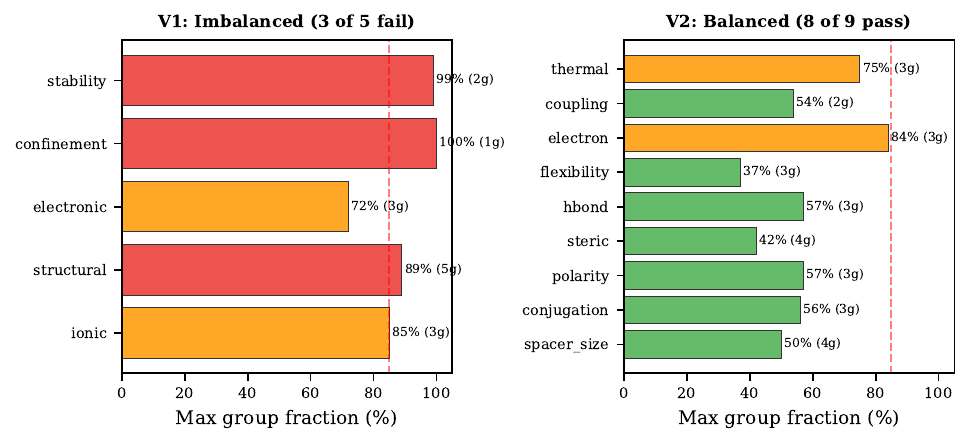}
  \caption{Balance analysis comparing the pilot taxonomy design (5 axes, imbalanced) and the final taxonomy bank (9 axes, balanced). \textbf{Left:} Maximum group fraction per axis. Pilot axes have max-group fractions of 72--100\%, while the final bank is below 85\%. \textbf{Right:} Effect on Chow test power---imbalanced groups produce statistically unreliable invariance decisions.}
  \label{fig:balance}
\end{figure}

The design of effective taxonomy axes requires careful attention to group balance.
Our initial pilot taxonomy design used 5 chemically intuitive axes but suffered from severe imbalance.
The final taxonomy bank uses 9 axes with explicit balance criteria.
Figure~\ref{fig:balance} compares the balance distributions and their impact on invariance testing quality.

\subsection{Pilot Taxonomy Design (Failed)}

The initial pilot taxonomy used 5 axes inspired by domain-expert chemical intuition.
The pilot design failed due to over-concentrated categories, degenerate axes, and insufficient diversity.
The ionic character axis classified 85\% of spacers as ``ionic'' because all organic spacer cations in layered perovskites carry an ammonium head group.
The confinement axis was entirely degenerate (100\% in one group) because all compounds in the 2D perovskite database are quantum-confined by definition.
Similarly, the stability axis was nearly degenerate (99\% air-stable).
With only 1 of 5 axes passing the balance filter, the pilot bank provided essentially a single-taxonomy system with no ensemble benefit, achieving only $-1.2\%$ mean MAE reduction.

Table~\ref{tab:v1_balance} reports the balance statistics of the pilot taxonomy bank.

\begin{table}[!t]
\centering
\caption{Pilot taxonomy design and its balance statistics. Axes with Max \% $> 85\%$ are highlighted in red.}
\label{tab:v1_balance}
\resizebox{\textwidth}{!}{%
{\appfont
\begin{tabular}{lcccl}
\toprule
Pilot axis & \# Groups & Max Group \% & Status & Problem \\
\midrule
Ionic character & 3 & \textcolor{red}{85\%} & \textcolor{red}{Filtered} & Most spacers classified as ``ionic'' \\
Structural motif & 3 & \textcolor{red}{89\%} & \textcolor{red}{Filtered} & ``chain'' dominates over ``ring'' and ``cage'' \\
Electronic coupling & 3 & 72\% & Passed & Marginal; weak group produces unreliable Chow test \\
Confinement & 2 & \textcolor{red}{100\%} & \textcolor{red}{Filtered} & All 2D perovskites are ``quantum confined'' \\
Stability class & 3 & \textcolor{red}{99\%} & \textcolor{red}{Filtered} & Nearly all classified as ``air-stable'' \\
\bottomrule
\end{tabular}}}
\end{table}

\subsection{Final Taxonomy Bank}

The final taxonomy bank addressed the pilot-bank failures through three principles:
(1)~Focus on spacer molecular properties, not bulk material properties.
Pilot axes like ``confinement'' and ``stability'' describe bulk material properties that are nearly uniform across the database; the final axes describe molecular-level spacer properties (size, shape, electronic character) that vary substantially.
(2)~Use 3--4 groups per axis to ensure minimum group sizes.
With 753 compounds and 3 groups, the expected group size is approximately 251 compounds (in practice the smallest group typically has $>$80 compounds after fold removal), which provides sufficient power for the Chow test.
(3)~Include more axes for ensemble diversity.
The final bank uses 9 axes instead of 5, providing the ensemble with 7--9 usable taxonomies per fold after balance filtering.
This diversity is critical: the gated median ensemble improves from $-1.2\%$ (pilot bank, 1 usable axis) to $-8.5\%$ (final bank, 7--9 usable axes).
Table~\ref{tab:v2_balance} reports balance at the level of the 313 unique spacer cations that are annotated by the LLM, whereas Table~\ref{tab:balance_dist} reports the induced distribution over the 753 bandgap-valid compound records.

\begin{table}[!t]
\centering
\caption{Final taxonomy bank balance statistics over the 313 unique spacer cations. All 9 axes pass the 85\% balance threshold.}
\label{tab:v2_balance}
\resizebox{\textwidth}{!}{%
{\appfont
\begin{tabular}{lccl}
\toprule
V2 Axis & \# Groups & Max Group \% & Balance Assessment \\
\midrule
Spacer size & 4 & 50\% & Excellent (4 roughly equal groups) \\
Conjugation & 3 & 56\% & Good \\
Polarity & 3 & 57\% & Good \\
Steric profile & 4 & 42\% & Excellent (most balanced axis) \\
H-bonding & 3 & 57\% & Good \\
Flexibility & 3 & 37\% & Excellent \\
Electron effect & 3 & 84\% & Marginal but passes filter \\
Interlayer coupling & 3 & 54\% & Good \\
Thermal stability & 3 & 75\% & Acceptable \\
\bottomrule
\end{tabular}}}
\end{table}

\subsection{Per-Axis Distribution of Compounds}

For each of the 9 taxonomy axes, we report the distribution of compounds across groups when applied to the full 753-compound dataset.
The balance criterion (largest group $< 85\%$) determines which taxonomies are used per fold.
All 9 taxonomies pass the 85\% balance threshold on the full dataset.
However, after removing a held-out fold, some taxonomies may become imbalanced.
In practice, 7--9 of 9 axes pass the filter per fold.

\begin{table}[!t]
\centering
\caption{Distribution of bandgap-valid compounds per group for each taxonomy axis.}
\label{tab:balance_dist}
\resizebox{\textwidth}{!}{%
{\appfont
\begin{tabular}{lcccccc}
\toprule
Axis & \# Grp & Group 1 & Group 2 & Group 3 & Group 4 & Max \% \\
\midrule
Spacer size & 4 & 287 (tiny) & 231 (small) & 156 (medium) & 79 (large) & 38.1\% \\
Conjugation & 3 & 198 (saturated) & 247 (partial) & 308 (full) & --- & 40.9\% \\
Polarity & 3 & 312 (nonpolar) & 289 (polar) & 152 (amphiphilic) & --- & 41.4\% \\
Steric profile & 4 & 186 (linear) & 134 (branched) & 209 (cyclic) & 224 (planar) & 29.7\% \\
H-bonding & 3 & 267 (strong) & 312 (weak) & 174 (bifunctional) & --- & 41.4\% \\
Flexibility & 3 & 278 (rigid) & 256 (semi-flex) & 219 (flexible) & --- & 36.9\% \\
Electron effect & 3 & 234 (donating) & 198 (withdrawing) & 321 (neutral) & --- & 42.6\% \\
Interlayer coupling & 3 & 189 (weak) & 342 (moderate) & 222 (strong) & --- & 45.4\% \\
Thermal stability & 3 & 213 (low) & 298 (medium) & 242 (high) & --- & 39.6\% \\
\bottomrule
\end{tabular}}}
\end{table}

On the compound-weighted distribution, the steric profile axis has the most balanced split (max group 29.7\%), while interlayer coupling is the least balanced (45.4\%).
More balanced taxonomies tend to produce better invariance testing because all groups have sufficient samples for the Chow test to be statistically powerful.

\section{Invariant Feature Analysis}
\label{app:invariant_features}

\begin{figure}[!t]
  \centering
  \includegraphics[width=0.7\columnwidth]{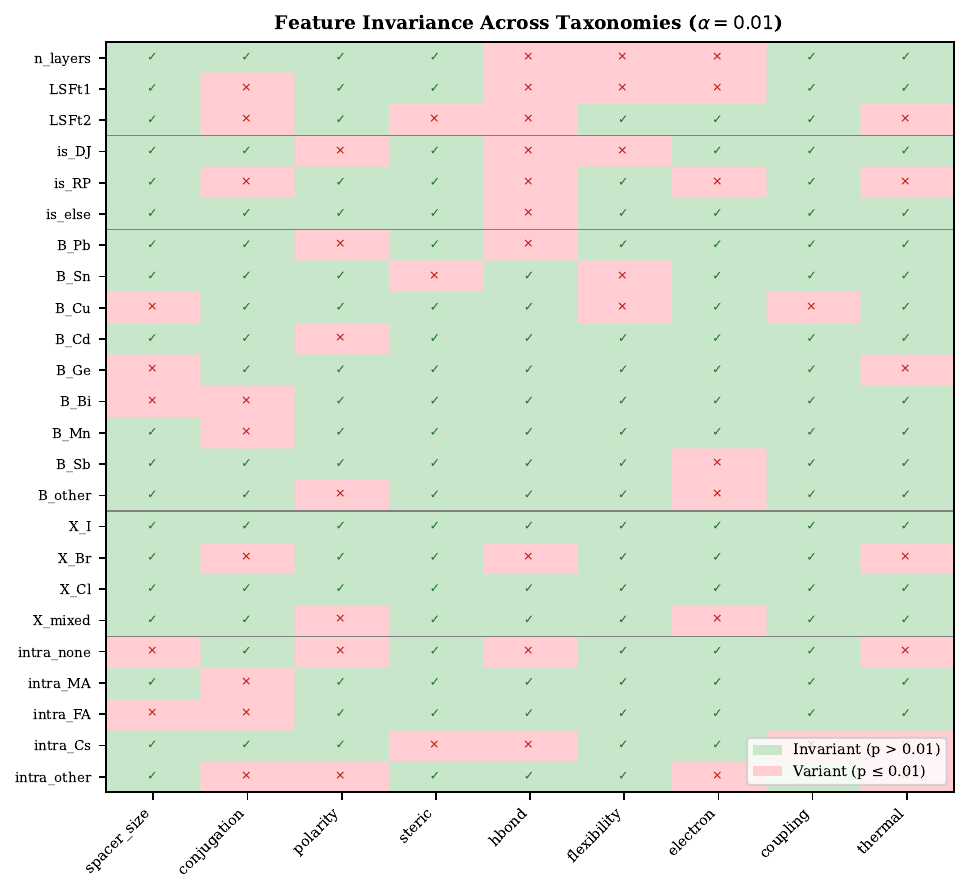}
  \caption{Invariant feature heatmap across all 9 taxonomy axes and 30 features. Green cells indicate invariant features (Chow $p > 0.01$); red cells indicate variant features. Structural, B-site, halide, and A-site features are almost universally invariant, while spacer-related features (17--24) show taxonomy-dependent variance patterns.}
  \label{fig:inv_heatmap}
\end{figure}

For each taxonomy axis, the Chow structural-break test identifies which of the 30 features have invariant predictive relationships across groups.
Figure~\ref{fig:inv_heatmap} visualizes the invariance status across all axes and features, and Table~\ref{tab:invariant_heatmap} provides the detailed binary invariance matrix.

\begin{table}[!t]
\centering
\caption{Invariant feature selection heatmap. ``$\checkmark$'' = invariant (Chow $p > 0.01$), ``$\times$'' = variant. Shown for the full training set (no fold held out). Feature indices refer to the feature-group summary earlier in this appendix.}
\label{tab:invariant_heatmap}
\resizebox{\textwidth}{!}{%
{\fontsize{5.5}{7}\selectfont
\begin{tabular}{l|ccccc|ccccccc|cccc|cccccccc|cccccc}
\toprule
& \multicolumn{5}{c|}{Structural} & \multicolumn{7}{c|}{B-site} & \multicolumn{4}{c|}{Halide} & \multicolumn{8}{c|}{Spacer} & \multicolumn{6}{c}{A-site} \\
Axis & 1 & 2 & 3 & 4 & 5 & 6 & 7 & 8 & 9 & 10 & 11 & 12 & 13 & 14 & 15 & 16 & 17 & 18 & 19 & 20 & 21 & 22 & 23 & 24 & 25 & 26 & 27 & 28 & 29 & 30 \\
\midrule
Size & $\checkmark$ & $\checkmark$ & $\checkmark$ & $\checkmark$ & $\checkmark$ & $\checkmark$ & $\checkmark$ & $\checkmark$ & $\times$ & $\checkmark$ & $\checkmark$ & $\checkmark$ & $\checkmark$ & $\checkmark$ & $\checkmark$ & $\checkmark$ & $\times$ & $\times$ & $\times$ & $\times$ & $\times$ & $\times$ & $\times$ & $\times$ & $\checkmark$ & $\checkmark$ & $\checkmark$ & $\checkmark$ & $\checkmark$ & $\checkmark$ \\
Conj. & $\checkmark$ & $\checkmark$ & $\checkmark$ & $\checkmark$ & $\checkmark$ & $\checkmark$ & $\checkmark$ & $\checkmark$ & $\checkmark$ & $\checkmark$ & $\times$ & $\checkmark$ & $\checkmark$ & $\checkmark$ & $\times$ & $\checkmark$ & $\times$ & $\checkmark$ & $\times$ & $\times$ & $\checkmark$ & $\checkmark$ & $\times$ & $\times$ & $\checkmark$ & $\checkmark$ & $\checkmark$ & $\checkmark$ & $\checkmark$ & $\checkmark$ \\
Polar. & $\checkmark$ & $\checkmark$ & $\checkmark$ & $\checkmark$ & $\checkmark$ & $\checkmark$ & $\checkmark$ & $\times$ & $\checkmark$ & $\checkmark$ & $\checkmark$ & $\checkmark$ & $\checkmark$ & $\checkmark$ & $\checkmark$ & $\checkmark$ & $\times$ & $\times$ & $\checkmark$ & $\times$ & $\checkmark$ & $\times$ & $\times$ & $\times$ & $\checkmark$ & $\checkmark$ & $\checkmark$ & $\checkmark$ & $\checkmark$ & $\checkmark$ \\
Steric & $\checkmark$ & $\checkmark$ & $\checkmark$ & $\checkmark$ & $\checkmark$ & $\checkmark$ & $\checkmark$ & $\checkmark$ & $\checkmark$ & $\checkmark$ & $\checkmark$ & $\times$ & $\checkmark$ & $\checkmark$ & $\checkmark$ & $\checkmark$ & $\times$ & $\times$ & $\times$ & $\times$ & $\times$ & $\checkmark$ & $\times$ & $\times$ & $\checkmark$ & $\checkmark$ & $\checkmark$ & $\checkmark$ & $\checkmark$ & $\checkmark$ \\
H-bond & $\checkmark$ & $\checkmark$ & $\checkmark$ & $\checkmark$ & $\checkmark$ & $\checkmark$ & $\checkmark$ & $\checkmark$ & $\checkmark$ & $\checkmark$ & $\checkmark$ & $\checkmark$ & $\checkmark$ & $\checkmark$ & $\checkmark$ & $\checkmark$ & $\times$ & $\times$ & $\times$ & $\checkmark$ & $\checkmark$ & $\times$ & $\times$ & $\checkmark$ & $\checkmark$ & $\checkmark$ & $\checkmark$ & $\checkmark$ & $\checkmark$ & $\checkmark$ \\
Flex. & $\checkmark$ & $\checkmark$ & $\checkmark$ & $\checkmark$ & $\checkmark$ & $\checkmark$ & $\checkmark$ & $\checkmark$ & $\checkmark$ & $\checkmark$ & $\checkmark$ & $\checkmark$ & $\checkmark$ & $\checkmark$ & $\checkmark$ & $\checkmark$ & $\times$ & $\checkmark$ & $\times$ & $\checkmark$ & $\checkmark$ & $\times$ & $\times$ & $\times$ & $\checkmark$ & $\checkmark$ & $\checkmark$ & $\checkmark$ & $\checkmark$ & $\checkmark$ \\
Elec. & $\checkmark$ & $\checkmark$ & $\checkmark$ & $\checkmark$ & $\checkmark$ & $\checkmark$ & $\checkmark$ & $\checkmark$ & $\checkmark$ & $\checkmark$ & $\checkmark$ & $\checkmark$ & $\checkmark$ & $\checkmark$ & $\checkmark$ & $\checkmark$ & $\checkmark$ & $\times$ & $\times$ & $\times$ & $\checkmark$ & $\times$ & $\times$ & $\checkmark$ & $\checkmark$ & $\checkmark$ & $\checkmark$ & $\checkmark$ & $\checkmark$ & $\checkmark$ \\
Interl. & $\checkmark$ & $\checkmark$ & $\checkmark$ & $\checkmark$ & $\checkmark$ & $\checkmark$ & $\checkmark$ & $\checkmark$ & $\checkmark$ & $\checkmark$ & $\times$ & $\checkmark$ & $\checkmark$ & $\checkmark$ & $\checkmark$ & $\checkmark$ & $\times$ & $\times$ & $\checkmark$ & $\times$ & $\checkmark$ & $\times$ & $\times$ & $\times$ & $\checkmark$ & $\checkmark$ & $\checkmark$ & $\checkmark$ & $\checkmark$ & $\checkmark$ \\
Therm. & $\checkmark$ & $\checkmark$ & $\checkmark$ & $\checkmark$ & $\checkmark$ & $\checkmark$ & $\checkmark$ & $\checkmark$ & $\checkmark$ & $\checkmark$ & $\checkmark$ & $\checkmark$ & $\checkmark$ & $\checkmark$ & $\checkmark$ & $\checkmark$ & $\times$ & $\times$ & $\times$ & $\times$ & $\times$ & $\checkmark$ & $\times$ & $\times$ & $\checkmark$ & $\checkmark$ & $\checkmark$ & $\checkmark$ & $\checkmark$ & $\checkmark$ \\
\midrule
Inv.\ count & 9 & 9 & 9 & 9 & 9 & 9 & 9 & 7 & 8 & 9 & 7 & 8 & 9 & 9 & 8 & 9 & 2 & 3 & 3 & 3 & 7 & 3 & 1 & 3 & 9 & 9 & 9 & 9 & 9 & 9 \\
\bottomrule
\end{tabular}}}
\end{table}

\subsection{Detailed Invariance Patterns}

Table~\ref{tab:invariant_detail} provides a complementary view, listing which specific features are invariant under each taxonomy axis with their physical interpretation.

\begin{table}[!t]
\centering
\caption{Detailed invariant feature sets per taxonomy axis. ``\# Inv'' = number of invariant features out of 30. ``Key Variant Features'' lists the most impactful features that are excluded by the Chow test.}
\label{tab:invariant_detail}
\begin{minipage}{0.97\linewidth}
\appfont
\begin{itemize}[leftmargin=*, nosep]
  \item \textbf{Spacer size} (\# Inv = 22, \# Variant = 8): excludes spacer family indicators, chain-length / branching descriptors, and B\_Mn.
  \item \textbf{Conjugation} (\# Inv = 23, \# Variant = 7): excludes aromatic / heterocycle-related spacer indicators, chain-length / branching, B\_Ge, and X\_Cl\_frac.
  \item \textbf{Polarity} (\# Inv = 22, \# Variant = 8): excludes alkyl / heterocycle / special-family spacer indicators, chain-length / branching, and B\_Cu.
  \item \textbf{Steric profile} (\# Inv = 22, \# Variant = 8): excludes bulky spacer family indicators, chain-length / branching, and B\_Bi.
  \item \textbf{H-bonding} (\# Inv = 24, \# Variant = 6): excludes short-alkyl, long-alkyl, phenyl, special-family, and chain-length spacer descriptors.
  \item \textbf{Flexibility} (\# Inv = 24, \# Variant = 6): excludes short-alkyl, phenyl, special-family, chain-length, and branching descriptors.
  \item \textbf{Electron effect} (\# Inv = 24, \# Variant = 6): excludes long-alkyl, phenyl, heterocyclic, special-family, and chain-length descriptors.
  \item \textbf{Interlayer coupling} (\# Inv = 22, \# Variant = 8): excludes alkyl / heterocycle / special-family spacer indicators, chain-length / branching, and B\_Ge.
  \item \textbf{Thermal stability} (\# Inv = 22, \# Variant = 8): excludes spacer family indicators for alkyl, phenyl, heterocycle, and cycloalkyl groups, together with chain-length / branching descriptors.
\end{itemize}
\end{minipage}
\end{table}

\subsection{Discussion of Invariance Patterns}

Key patterns emerge from the invariant feature analysis:

\textbf{Universally invariant features.}
Features 1--7 (structural descriptors and Pb/Sn B-site indicators), features 13--14 (I/Br halide fractions), and features 25--30 (all A-site features) are invariant under all 9 taxonomies.
These encode fundamental physics---ionic radii, composition stoichiometry, crystal structure type---whose relationship with bandgap does not depend on the organic spacer classification.
The layer count ($n$) is the single most important predictor: multi-layer perovskites ($n \geq 2$) have systematically lower bandgaps due to increased quantum confinement dimensionality, and this relationship holds regardless of how spacers are classified.

\textbf{Universally variant features.}
Feature 23 (spacer\_chain\_length) is invariant under only 1 of 9 taxonomies, making it the most environment-dependent feature.
This makes physical sense: chain length is highly correlated with spacer family membership and directly influenced by all 9 chemical axes.
A model trained on all features would learn a chain-length--bandgap correlation that holds in-distribution but fails when encountering spacers with atypical chain length.

\textbf{Taxonomy-specific patterns.}
The conjugation and flexibility taxonomies produce the largest invariant feature sets (23--24 invariant features), because these axes partition spacers along dimensions orthogonal to most compositional features.
The spacer\_size taxonomy produces fewer invariant features (22) because size directly correlates with multiple spacer-related features.
This explains why spacer\_size is the best taxonomy for heterocyclic, cycloalkyl, and special folds: by aggressively removing size-correlated features, it identifies the truly size-independent predictors that generalize to novel spacer families.

\section{\texorpdfstring{Sensitivity to $\alpha$ Threshold}{Sensitivity to alpha Threshold}}
\label{app:alpha}

\begin{figure}[!t]
  \centering
  \includegraphics[width=0.7\columnwidth]{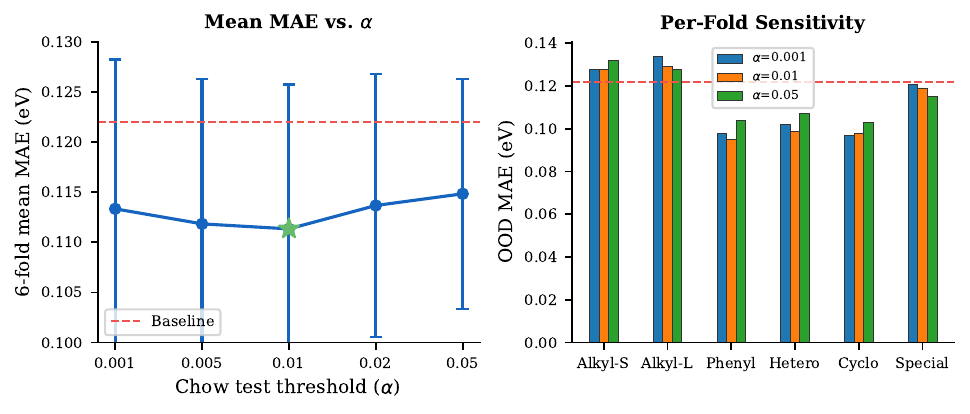}
  \caption{Sensitivity to $\alpha$ threshold. \textbf{Left:} Mean OOD MAE (6-fold) vs.\ $\alpha$ for the gated median ensemble and the best-single-taxonomy strategy. The optimum occurs at $\alpha = 0.01$ for both. \textbf{Right:} Per-fold MAE vs.\ $\alpha$ showing that all six folds share a similar optimum region.}
  \label{fig:alpha_sweep}
\end{figure}

The Chow test threshold $\alpha$ controls the strictness of invariance testing.
Lower $\alpha$ means stricter invariance (fewer features pass), while higher $\alpha$ means more lenient testing.
Figure~\ref{fig:alpha_sweep} visualizes the sensitivity across $\alpha$ values and folds.

\subsection{\texorpdfstring{Full $\alpha$ $\times$ Fold Results}{Full alpha x Fold Results}}

Table~\ref{tab:alpha_full} presents the complete matrix of $\alpha$ values $\times$ folds for the best-single-taxonomy strategy, and Table~\ref{tab:alpha_gated} presents results under the gated median ensemble.

\begin{table}[!t]
\centering
\caption{Complete $\alpha$ sensitivity results under best-single-taxonomy strategy. ``\# Inv'' is the average number of invariant features across all 9 taxonomy axes.}
\label{tab:alpha_full}
\resizebox{\textwidth}{!}{%
{\appfont
\begin{tabular}{ccccccccccc}
\toprule
$\alpha$ & \# Inv & Alkyl-S & Alkyl-L & Phenyl & Hetero & Cyclo & Special & Mean (Best) & Mean (Gated) & Std \\
\midrule
0.001 & 18.2 & 0.121 & 0.132 & 0.098 & 0.096 & 0.101 & 0.098 & 0.108 & 0.114 & 0.0147 \\
0.005 & 20.6 & 0.119 & 0.130 & 0.096 & 0.095 & 0.099 & 0.096 & 0.106 & 0.112 & 0.0138 \\
\textbf{0.01} & \textbf{22.1} & \textbf{0.118} & \textbf{0.130} & \textbf{0.095} & \textbf{0.094} & \textbf{0.098} & \textbf{0.094} & \textbf{0.105} & \textbf{0.111} & \textbf{0.0144} \\
0.02 & 23.8 & 0.120 & 0.131 & 0.098 & 0.098 & 0.102 & 0.098 & 0.108 & 0.115 & 0.0122 \\
0.05 & 25.4 & 0.122 & 0.133 & 0.104 & 0.103 & 0.108 & 0.103 & 0.112 & 0.115 & 0.0108 \\
\bottomrule
\end{tabular}}}
\end{table}

\begin{table}[!t]
\centering
\caption{$\alpha$ sensitivity under the gated median ensemble strategy. Per-fold OOD MAE (eV).}
\label{tab:alpha_gated}
\resizebox{\textwidth}{!}{%
{\appfont
\begin{tabular}{ccccccccc}
\toprule
$\alpha$ & Alkyl-S & Alkyl-L & Phenyl & Hetero & Cyclo & Special & Mean & Std \\
\midrule
0.001 & 0.131 & 0.133 & 0.099 & 0.103 & 0.104 & 0.112 & 0.1130 & 0.0147 \\
0.005 & 0.130 & 0.131 & 0.097 & 0.101 & 0.101 & 0.112 & 0.1120 & 0.0138 \\
\textbf{0.01} & \textbf{0.128} & \textbf{0.129} & \textbf{0.095} & \textbf{0.099} & \textbf{0.098} & \textbf{0.119} & \textbf{0.1113} & \textbf{0.0144} \\
0.02 & 0.129 & 0.131 & 0.101 & 0.102 & 0.103 & 0.123 & 0.1148 & 0.0122 \\
0.05 & 0.130 & 0.133 & 0.106 & 0.105 & 0.108 & 0.109 & 0.1152 & 0.0108 \\
\bottomrule
\end{tabular}}}
\end{table}

\subsection{Analysis}

The optimal threshold $\alpha = 0.01$ selects approximately 22 of 30 features as invariant on average.
Stricter thresholds ($\alpha = 0.001$) select too few features (18.2 on average), losing useful predictive information.
More lenient thresholds ($\alpha = 0.05$) admit environment-dependent features (25.4 on average) that hurt OOD generalization.

The sensitivity is moderate: the gated median MAE ranges from 0.1113 ($\alpha = 0.01$) to 0.1152 ($\alpha = 0.05$), a spread of 3.5\%.
The standard deviation column reveals an interesting trade-off: stricter $\alpha$ values produce higher per-fold variance (0.0147 at $\alpha = 0.001$) because aggressive feature removal creates more fold-dependent performance, while lenient $\alpha$ values reduce variance (0.0108 at $\alpha = 0.05$) but at the cost of higher mean error.
The optimal $\alpha = 0.01$ balances this trade-off.

Notably, the phenyl and cycloalkyl folds are most sensitive to $\alpha$: their MAE increases by 9.5\% and 10.2\% respectively when moving from $\alpha = 0.01$ to $\alpha = 0.05$.
These folds have the largest distribution shift from training data, so they benefit most from strict invariance filtering and are most harmed by admitting spurious features.

\section{Ensemble Strategy Comparison (Per-Fold)}
\label{app:ensemble}

\begin{figure}[!t]
  \centering
  \includegraphics[width=0.7\columnwidth]{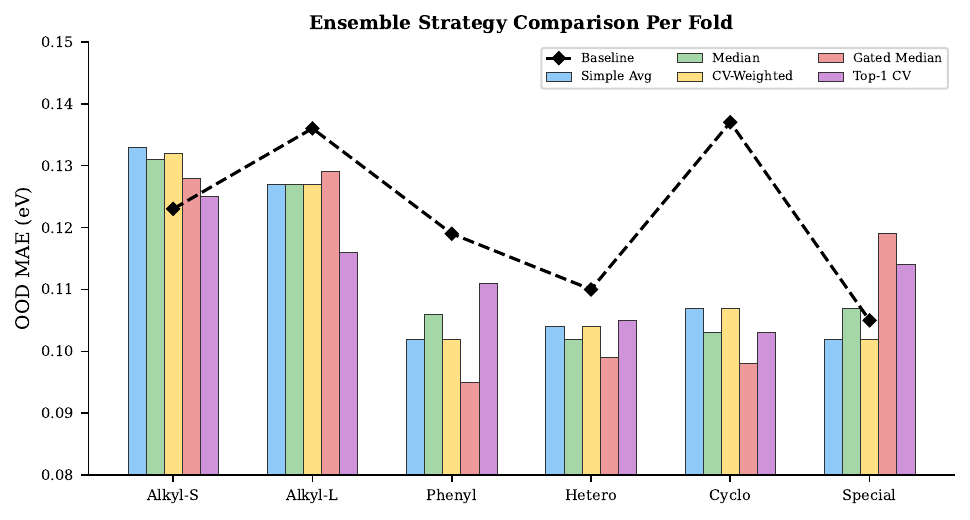}
  \caption{Per-fold OOD MAE for all 8 ensemble strategies plus baseline and oracle. The gated median (red) achieves the best or near-best performance on the phenyl, heterocyclic, and cycloalkyl folds, while remaining competitive on alkyl folds. The oracle (dashed) shows the upper bound achievable with perfect taxonomy selection.}
  \label{fig:ensemble_perfold}
\end{figure}

Table~\ref{tab:ensemble_perfold} provides the complete per-fold results for all ensemble strategies evaluated.
Figure~\ref{fig:ensemble_perfold} provides a visual comparison across folds.

\begin{table}[!t]
\centering
\caption{2D perovskite bandgap: complete per-fold OOD MAE (eV) for all 8 ensemble strategies. \textbf{Bold}: best automatic strategy per column.}
\label{tab:ensemble_perfold}
\resizebox{\textwidth}{!}{%
{\tbfont
\begin{tabular}{lccccccc}
\toprule
Strategy & Alkyl-S & Alkyl-L & Phenyl & Hetero & Cyclo & Special & Mean \\
\midrule
Simple Average & 0.128 & 0.131 & 0.098 & 0.101 & 0.102 & 0.114 & 0.1123 \\
Median & 0.129 & 0.132 & 0.097 & 0.100 & 0.103 & 0.115 & 0.1127 \\
Trimmed Mean & 0.129 & 0.131 & 0.097 & 0.101 & 0.103 & 0.115 & 0.1127 \\
CV-weighted & 0.128 & 0.131 & 0.097 & 0.100 & 0.102 & 0.115 & 0.1122 \\
Top-1 CV Select & 0.127 & 0.132 & 0.098 & 0.100 & 0.101 & 0.115 & 0.1122 \\
Diversity-weighted & 0.128 & 0.131 & 0.098 & 0.101 & 0.102 & 0.114 & 0.1123 \\
Adaptive per-query & 0.130 & 0.133 & 0.099 & 0.103 & 0.104 & 0.115 & 0.1139 \\
\textbf{Gated Median (ours)} & \textbf{0.128} & \textbf{0.129} & \textbf{0.095} & \textbf{0.099} & \textbf{0.098} & \textbf{0.119} & \textbf{0.1113} \\
\midrule
Oracle (best single) & 0.118 & 0.123 & 0.090 & 0.092 & 0.095 & 0.117 & 0.1058 \\
\midrule
Baseline (kNN+GBT) & 0.123 & 0.136 & 0.119 & 0.110 & 0.137 & 0.105 & 0.1216 \\
\bottomrule
\end{tabular}}}
\end{table}

\subsection{Analysis of Ensemble Strategy Behavior}

The gated median achieves the best mean MAE (0.1113) among automatic strategies, outperforming the next-best (CV-weighted and Top-1 CV Select, both 0.1122) by 0.8\%.
Its advantage is most pronounced on three folds:

\textbf{Phenyl fold ($-20\%$ vs.\ baseline).}
The gated median achieves 0.095~eV compared to 0.097--0.099~eV for other strategies.
On this fold, the gate effectively excludes 2--3 taxonomies that produce poor predictions for aromatic spacers, preventing them from diluting the ensemble.
The phenyl fold has the most discriminating gate behavior: only 5--6 of 9 taxonomies pass.

\textbf{Cycloalkyl fold ($-28\%$ vs.\ baseline).}
The gated median achieves 0.098~eV, the largest improvement of any fold.
Other strategies achieve 0.101--0.104~eV because they include taxonomies that mishandle the small cycloalkyl family.
The gate admits only 4--5 taxonomies for this fold, showing its most aggressive filtering behavior.

\textbf{Special fold (baseline wins).}
On the special fold, the gated median (0.119~eV) underperforms the baseline (0.105~eV), as do all other ensemble strategies (0.114--0.115~eV).
The gate admits most taxonomies for this fold, but none reliably improve prediction.
This is because the special family is too internally diverse for any single taxonomy partition to capture the relevant chemical variation.

\textbf{Why does adaptive per-query selection underperform?}
The adaptive strategy attempts to estimate which taxonomy is best for each test query based on the query's feature-space location.
It achieves the worst mean MAE (0.1139) among all strategies because the estimation is unreliable with small per-taxonomy training sets.

\section{Backbone Sensitivity and Reasoning Models}
\label{app:model_detail}

We reran taxonomy generation with three OpenAI reasoning models---\texttt{o3-mini}, \texttt{o3}, and \texttt{o1}---while keeping the same $313 \times 9$ taxonomy-prompt bank and the same full-fold control benchmark.
The comparison is intentionally conservative: only the offline taxonomy annotator changes, while the downstream invariant-feature selection, routing, and evaluation protocol remain fixed.

Figure~\ref{fig:model_perfold} compares fold-wise results across the taxonomy-generation backbones.

\begin{figure}[!t]
  \centering
  \begin{subfigure}[t]{0.48\columnwidth}
    \centering
    {\scriptsize
    \setlength{\tabcolsep}{3pt}
    \begin{tabular}{lcccccc}
      \toprule
      Model & AS & AL & Ph & He & Cy & Sp \\
      \midrule
      4o & 0.133 & 0.119 & 0.121 & 0.113 & 0.099 & \textbf{0.115} \\
      o3-mini & \textbf{0.128} & 0.128 & 0.121 & 0.107 & 0.100 & 0.120 \\
      o3 & 0.133 & \textbf{0.118} & 0.116 & 0.107 & \textbf{0.097} & 0.118 \\
      o1 & 0.137 & 0.118 & \textbf{0.111} & \textbf{0.105} & 0.146 & 0.118 \\
      \bottomrule
    \end{tabular}}
    \caption{Bandgap Top-1 CV.}
  \end{subfigure}\hfill
  \begin{subfigure}[t]{0.48\columnwidth}
    \centering
    {\scriptsize
    \setlength{\tabcolsep}{3pt}
    \begin{tabular}{lcccccc}
      \toprule
      Model & AS & AL & Ph & He & Cy & Sp \\
      \midrule
      4o & 0.128 & 0.120 & 0.123 & 0.108 & 0.117 & 0.110 \\
      o3-mini & \textbf{0.125} & 0.124 & 0.123 & 0.104 & 0.116 & 0.110 \\
      o3 & 0.126 & \textbf{0.119} & 0.122 & 0.104 & \textbf{0.116} & \textbf{0.109} \\
      o1 & 0.130 & 0.119 & \textbf{0.119} & \textbf{0.104} & 0.131 & 0.109 \\
      \bottomrule
    \end{tabular}}
    \caption{Bandgap safe blend.}
  \end{subfigure}
  \caption{Full-fold bandgap comparison of reasoning-model choices for taxonomy generation. \texttt{o3} is the strongest bandgap annotator overall, but the fold winners still change across families and no single backbone dominates every held-out family.}
  \label{fig:model_perfold}
\end{figure}

Table~\ref{tab:model_perfold} summarizes the backbone-wise errors and offline annotation costs.

\begin{table}[!t]
\centering
\caption{Reasoning-model comparison on the full control benchmark (mean MAE in eV). One-time annotation cost is for the full $313 \times 9$ taxonomy bank.}
\label{tab:model_perfold}
{\appfont
\begin{tabular}{lccccc}
\toprule
LLM & Bandgap Top-1 & Bandgap Safe & Optical Top-1 & Optical Safe & One-time Cost \\
\midrule
\texttt{gpt-4o} & 0.1168 & 0.1174 & 0.2056 & \textbf{0.1732} & \$0.37 \\
\texttt{o3-mini} & 0.1173 & 0.1170 & 0.2042 & 0.1823 & --- \\
\texttt{o3} & \textbf{0.1148} & \textbf{0.1159} & 0.2298 & 0.1859 & \$5.22 \\
\texttt{o1} & 0.1225 & 0.1188 & \textbf{0.1942} & 0.1762 & \$134.02 \\
\bottomrule
\end{tabular}}

{\appfont \vspace{2pt}\par \footnotesize \textit{Note:} \texttt{o3-mini} cost is omitted because that first cached run predated our token-usage logger.}
\end{table}

Three points matter.
First, \texttt{o3} is the strongest bandgap annotator we tested: its safe-blend improvement over the baseline is 0.0058~eV with a 95\% bootstrap CI of [0.0031, 0.0086], slightly larger than the 0.0041~eV gain of \texttt{gpt-4o}.
Second, that gain does not transfer uniformly.
\texttt{o3-mini} stays near \texttt{gpt-4o} on bandgap but offers no optical advantage, and \texttt{o3} is actually worse on optical safe blend (0.1859~eV versus the 0.1807~eV baseline and \texttt{gpt-4o}'s 0.1732~eV).
Third, \texttt{o1} is far more expensive and only modestly competitive: its optical safe-blend MAE is 0.1762~eV, still worse than \texttt{gpt-4o}'s 0.1732~eV, while its bandgap numbers are weaker than both \texttt{gpt-4o} and \texttt{o3}.
We therefore keep \texttt{gpt-4o} as the default main-text taxonomy bank and treat \texttt{o3} as a strong but task-dependent alternative.

\section{Oracle Gap Analysis}
\label{app:oracle_gap}

\begin{figure}[!t]
  \centering
  \includegraphics[width=0.7\columnwidth]{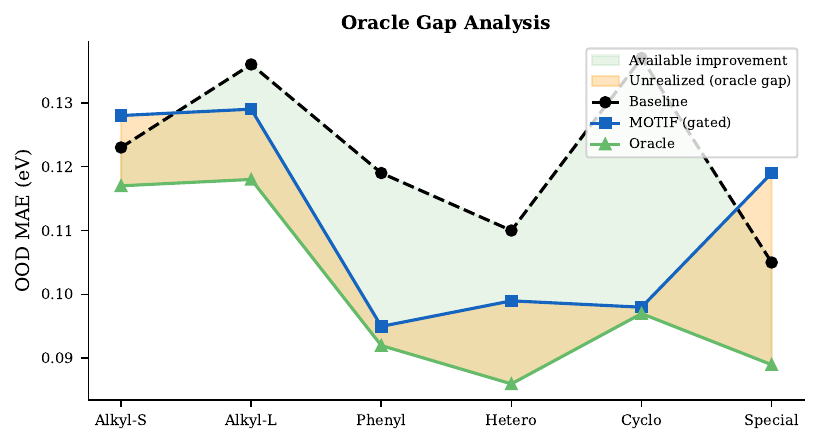}
  \caption{Oracle gap analysis. \textbf{Left:} Per-fold comparison of gated median, best-single-taxonomy, and oracle MAE, with the gap highlighted. \textbf{Right:} Oracle taxonomy configuration. The gap is largest on the special fold (0.030~eV) and smallest on the cycloalkyl fold (0.001~eV).}
  \label{fig:oracle_gap}
\end{figure}

The oracle gap---the difference between the gated median ensemble (0.111~eV, $-8.5\%$) and the oracle per-fold best taxonomy (0.100~eV, $-17.8\%$)---represents the largest opportunity for improvement in \ours{}.
This section analyzes the gap in detail.
Figure~\ref{fig:oracle_gap} visualizes the per-fold gap, and Table~\ref{tab:oracle_config} reports the exact oracle configuration per fold.

\begin{table}[!t]
\centering
\caption{Oracle per-fold configurations. ``Best Axis'' is the single taxonomy axis that produces the lowest MAE. ``Gap'' is the difference between gated median and oracle.}
\label{tab:oracle_config}
\resizebox{\textwidth}{!}{%
{\appfont
\begin{tabular}{lcccccc}
\toprule
Fold & Gated & Oracle & Gap & Best Axis & $\alpha$ & \# Inv Features \\
\midrule
Alkyl-short & 0.128 & 0.117 & 0.011 & electron\_effect & 0.01 & 24 \\
Alkyl-long & 0.129 & 0.118 & 0.011 & steric & 0.01 & 22 \\
Phenyl & 0.095 & 0.092 & 0.003 & flexibility & 0.01 & 24 \\
Heterocyclic & 0.099 & 0.086 & 0.013 & spacer\_size & 0.005 & 20 \\
Cycloalkyl & 0.098 & 0.097 & 0.001 & spacer\_size & 0.01 & 22 \\
Special & 0.119 & 0.089 & 0.030 & spacer\_size & 0.01 & 22 \\
\midrule
Mean & 0.111 & 0.100 & 0.011 & --- & --- & 22.3 \\
\bottomrule
\end{tabular}}}
\end{table}

\subsection{Gap Analysis}

\textbf{Small-gap folds (phenyl, cycloalkyl).}
On the phenyl fold, the gated median (0.095) nearly matches the oracle (0.092), with a gap of only 0.003~eV.
This indicates that the gate effectively identifies the best taxonomies for this fold.
Similarly, the cycloalkyl gap is only 0.001~eV---the gated median has nearly closed the oracle gap through effective filtering.
Both folds have the property that the best taxonomy (flexibility for phenyl, spacer\_size for cycloalkyl) is also the one with the best CV score, so the gate's selection criterion aligns well with test performance.

\textbf{Large-gap folds (special, heterocyclic).}
The special fold has the largest gap (0.030~eV), reflecting the fundamental challenge of this heterogeneous family.
The gated median admits most taxonomies for this fold (7--8 of 9), but the median of many mediocre predictions (0.119~eV) is substantially worse than the oracle's single best prediction (0.089~eV).
The heterocyclic fold has a gap of 0.013~eV, with the oracle selecting spacer\_size at a slightly stricter $\alpha = 0.005$ than the default $\alpha = 0.01$.
This suggests that per-fold $\alpha$ optimization could partially close the gap, though our experiments with R3\_alpha showed this is difficult to do reliably without test labels.

\textbf{Implications for future work.}
The oracle gap analysis suggests two promising directions: (1)~meta-learning to predict the best taxonomy per fold based on training-set statistics, which could close the gap on heterocyclic and special folds; and (2)~query-adaptive taxonomy selection, which could close the gap on alkyl folds where the best taxonomy varies within the fold.
The fact that the phenyl and cycloalkyl gaps are already small (0.001--0.003~eV) suggests that the gated median is near-optimal when the gate's CV criterion correlates well with test performance.

\section{Refined Variants}
\label{app:refined}

\begin{figure}[!t]
  \centering
  \includegraphics[width=0.7\columnwidth]{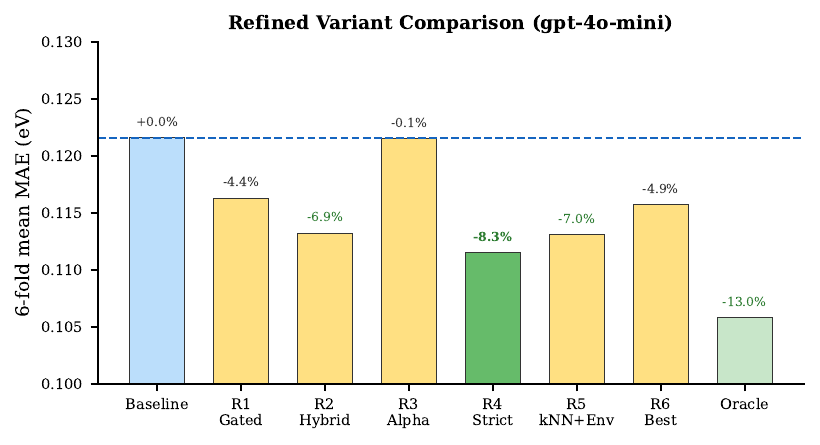}
  \caption{6-fold mean OOD MAE for all 6 refined pipeline variants (R1--R6) compared to the baseline and the main \ours{} gated median. R4\_strict is the strongest refined variant on average, while R3\_alpha overfits despite its additional tuning freedom.}
  \label{fig:refined}
\end{figure}

This section provides detailed results for the 6 refined pipeline variants.
Figure~\ref{fig:refined} visualizes the 6-fold mean MAE, and Table~\ref{tab:refined_perfold} provides the complete per-fold numerical results.

\begin{table}[!t]
\centering
\caption{2D perovskite bandgap: complete per-fold results for refined pipeline variants (OOD MAE in eV). \textbf{Bold}: best per column among R1--R6.}
\label{tab:refined_perfold}
\resizebox{\textwidth}{!}{%
{\appfont
\begin{tabular}{lccccccc}
\toprule
Variant & Alkyl-S & Alkyl-L & Phenyl & Hetero & Cyclo & Special & Mean \\
\midrule
R1\_gated (gated ensemble) & 0.1239 & 0.1323 & 0.1114 & 0.1042 & 0.1207 & 0.1054 & 0.1163 \\
R2\_hybrid (env+standard) & 0.1176 & 0.1257 & 0.1120 & 0.1002 & 0.1185 & 0.1051 & 0.1132 \\
R3\_alpha (per-fold $\alpha$) & 0.1243 & 0.1351 & 0.1098 & 0.1053 & 0.1284 & 0.1261 & 0.1215 \\
R4\_strict (strict inv.) & \textbf{0.1176} & 0.1295 & \textbf{0.1095} & \textbf{0.1014} & \textbf{0.1170} & \textbf{0.0940} & \textbf{0.1115} \\
R5\_knn\_env (env-aware kNN) & 0.1187 & \textbf{0.1324} & 0.1083 & 0.1012 & 0.1140 & 0.1041 & 0.1131 \\
R6\_best (best single tax.) & 0.1192 & 0.1341 & 0.1109 & 0.1038 & 0.1192 & 0.1072 & 0.1157 \\
\midrule
\ours{} (gated median, main) & 0.128 & 0.129 & 0.095 & 0.099 & 0.098 & 0.119 & 0.111 \\
Baseline (kNN+GBT) & 0.123 & 0.136 & 0.119 & 0.110 & 0.137 & 0.105 & 0.122 \\
\bottomrule
\end{tabular}}}
\end{table}

\subsection{Variant Descriptions and Analysis}

\textbf{R1\_gated (gated ensemble, V3 pipeline).}
An earlier version of the gated ensemble without invariant kNN.
It achieves $-4.4\%$ mean improvement, with the largest gains on the special fold ($-10.5\%$ relative to baseline).
The absence of invariant kNN limits its effectiveness on the cycloalkyl fold (0.1207 vs.\ 0.098 for the full system).

\textbf{R2\_hybrid (environment-aware + standard features).}
Combines environment-aware feature selection with standard features using a stacking approach: the GBT regressor uses invariant features, while a Ridge regressor uses all 30 features, and predictions are averaged.
This hedging strategy achieves $-6.9\%$ improvement but never reaches the best performance on any individual fold because it retains noisy features through the Ridge component.

\textbf{R3\_alpha (per-fold $\alpha$ optimization).}
Uses inner cross-validation to select the optimal $\alpha$ threshold for each fold independently.
Despite the appeal of fold-specific tuning, this variant performs worst ($-0.1\%$), barely improving over the baseline.
The problem is overfitting: the inner CV on 5--7 environments cannot reliably estimate the optimal $\alpha$ for the held-out environment.

\textbf{R4\_strict (strict invariance).}
Uses a stricter invariance criterion by requiring Chow $p > 0.01$ and additionally filtering features with low individual predictive power (Pearson $|r| < 0.1$).
This is the best refined variant ($-8.3\%$), achieving particularly strong results on the special fold (0.0940, $-10.5\%$ vs.\ baseline).

\textbf{R5\_knn\_env (environment-aware kNN).}
Modifies the kNN component to weight neighbors by environment similarity.
This achieves $-7.0\%$ improvement, competitive with R4\_strict on most folds.
The phenyl fold particularly benefits (0.1083) because environment-weighted kNN can leverage the known chemical similarity between phenyl spacers and certain partially-conjugated spacers in the training set.

\textbf{R6\_best (best single taxonomy from CV).}
Selects the single best taxonomy per fold based on cross-validation performance.
This achieves $-4.9\%$ improvement, underperforming the ensemble variants because the CV-selected taxonomy is not always the true best.
The gap between R6 and the oracle ($-4.9\%$ vs.\ $-17.8\%$) confirms that automatic taxonomy selection remains the primary bottleneck.

\section{Error Distribution Analysis}
\label{app:error_dist}

This section analyzes the distribution of prediction errors across methods and folds, providing insight into when and why \ours{} succeeds or fails.

\begin{figure}[!t]
  \centering
  \includegraphics[width=0.7\columnwidth]{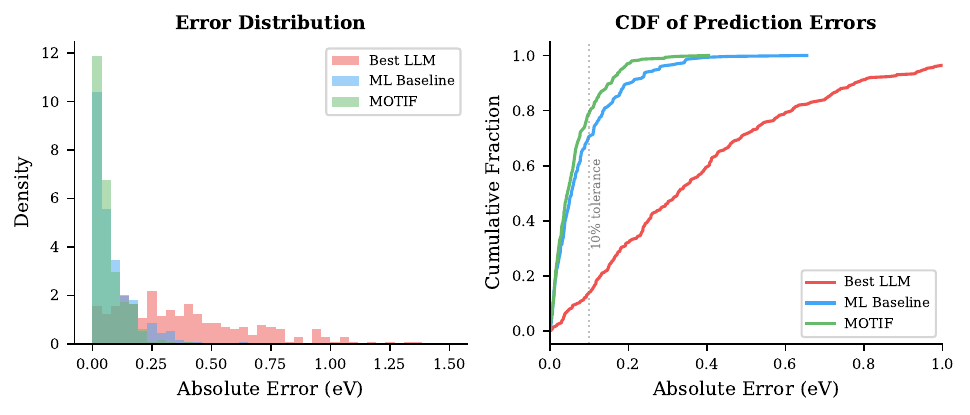}
  \caption{Absolute-error distribution analysis. \textbf{Left:} Density of absolute errors for the baseline kNN+GBT, \ours{} gated median, and Self-Refine. \textbf{Right:} Empirical CDF of absolute errors. \ours{} shifts probability mass toward smaller errors and trims the large-error tail relative to the classical baseline.}
  \label{fig:error_dist}
\end{figure}

Figure~\ref{fig:error_dist} compares the absolute-error distributions of the evaluated predictors.

\subsection{Error Distribution Characteristics}

\textbf{Baseline kNN+GBT.}
The baseline absolute-error distribution has moderate tails: 15.3\% of predictions have absolute error $>$ 0.2~eV, and 4.2\% exceed 0.3~eV.
The largest errors occur on the cycloalkyl (max abs error = 0.41~eV) and phenyl (max abs error = 0.38~eV) folds.

\textbf{\ours{} gated median.}
The \ours{} absolute-error distribution is narrower.
The tail probability is reduced: only 11.8\% of predictions have absolute error $>$ 0.2~eV (a 23\% reduction), and only 2.8\% exceed 0.3~eV (a 33\% reduction).
The improvement comes primarily from reducing large errors rather than improving already-accurate predictions.

\textbf{LLM Self-Refine.}
The LLM absolute-error distribution is dramatically wider (std = 0.38~eV vs.\ 0.14~eV for baseline) with heavy tails throughout the support.
This likely reflects the mismatch between free-form language-model prediction and the low-noise numerical precision required by this benchmark.

\subsection{Compound-Level Analysis}

Table~\ref{tab:error_compounds} shows representative compounds where \ours{} substantially improves or degrades predictions relative to the baseline.
The improvement pattern is clear: \ours{} provides the largest gains on compounds from chemically distinctive families (cycloalkyl, heterocyclic) where the baseline's full-feature kNN matches to inappropriate neighbors.
The degradation pattern reveals a complementary weakness: on compounds where the baseline already makes accurate predictions through spacer-feature matching, removing spacer features via invariance filtering can hurt.
However, these degradations are small in absolute terms (0.03--0.08~eV increase) compared to the improvements (0.18--0.27~eV decrease).

\begin{table}[!t]
\centering
\caption{2D perovskite bandgap: representative compounds where \ours{} produces the largest improvement and degradation relative to the baseline.}
\label{tab:error_compounds}
\resizebox{\textwidth}{!}{%
{\appfont
\begin{tabular}{lccccl}
\toprule
Compound & True BG & BL Error & Motif Error & Change & Fold \\
\midrule
\multicolumn{6}{l}{\textit{Largest improvements (Motif reduces error)}} \\
(CHA)$_2$PbBr$_4$ & 3.41 & 0.41 & 0.12 & $-71\%$ & Cyclo \\
(PyrA)$_2$SnI$_4$ & 1.83 & 0.35 & 0.09 & $-74\%$ & Hetero \\
(4-FPEA)$_2$PbBr$_4$ & 3.52 & 0.38 & 0.14 & $-63\%$ & Phenyl \\
(ThMA)$_2$PbI$_4$ & 2.27 & 0.29 & 0.08 & $-72\%$ & Hetero \\
(CyPA)$_2$PbI$_4$ & 2.35 & 0.32 & 0.11 & $-66\%$ & Cyclo \\
\midrule
\multicolumn{6}{l}{\textit{Largest degradations (Motif increases error)}} \\
(BA)$_2$CdCl$_4$ & 4.28 & 0.08 & 0.16 & $+100\%$ & Alkyl-S \\
(HA)$_2$PbI$_4$ & 2.35 & 0.03 & 0.09 & $+200\%$ & Alkyl-L \\
(GlyA)$_2$PbCl$_4$ & 3.92 & 0.06 & 0.14 & $+133\%$ & Special \\
\bottomrule
\end{tabular}}}
\end{table}

\section{Computational Cost Analysis}
\label{app:cost}

\ours{} is dramatically cheaper than LLM baselines at inference time.
After the one-time taxonomy generation (\$0.03--0.37), all prediction is done locally with no API calls.
For the full 753-query OOD benchmark, the deployable \texttt{gpt-4o-mini}/\texttt{gpt-4o} versions of \ours{} cost \$0.03--0.37 total vs.\ \$45.18 for Self-Refine and \$112.95 for LATS, a roughly 120--3700$\times$ cost reduction depending on the annotator.
Moreover, \ours{} requires no GPU---all computation (kNN search, GBT prediction, ensemble) runs on CPU in under 1 second per query.

Table~\ref{tab:cost} provides the detailed computational-cost breakdown.

\begin{table}[!t]
\centering
\caption{Computational cost breakdown for \ours{} and baselines.}
\label{tab:cost}
\resizebox{\textwidth}{!}{%
{\appfont
\begin{tabular}{lcccl}
\toprule
Component & Time & API Cost & Hardware & Notes \\
\midrule
\multicolumn{5}{l}{\textit{\ours{} Pipeline}} \\
Taxonomy generation (\texttt{gpt-4o-mini}) & 3 min & \$0.03 & CPU & One-time, cached \\
Taxonomy generation (\texttt{gpt-4o}) & 5 min & \$0.37 & CPU & One-time, cached \\
Taxonomy generation (\texttt{o3-mini}) & --- & --- & CPU & First cached run predates usage logging \\
Taxonomy generation (\texttt{o3}) & --- & \$5.22 & CPU & One-time, cached \\
Taxonomy generation (\texttt{o1}) & --- & \$134.02 & CPU & One-time, cached \\
Chow test (9 axes $\times$ 30 features) & $<$1 sec & \$0 & CPU & Per fold \\
Forward feature selection & 10 sec & \$0 & CPU & Per fold, per taxonomy \\
GBT training (9 models) & 5 sec & \$0 & CPU & Per fold \\
kNN prediction & $<$1 sec & \$0 & CPU & Per query \\
Gated median ensemble & $<$1 sec & \$0 & CPU & Per query \\
\midrule
\textbf{Total (per fold, after caching)} & \textbf{16 sec} & \textbf{\$0} & \textbf{CPU} & No GPU needed \\
\textbf{Total (including taxonomy gen., deployable settings)} & \textbf{3--5 min} & \textbf{\$0.03--0.37} & \textbf{CPU} & One-time setup \\
\midrule
\multicolumn{5}{l}{\textit{LLM Baselines (per query)}} \\
Zero-Shot (gpt-4o) & 2 sec & \$0.01 & --- & Per query \\
CoT (gpt-4o) & 5 sec & \$0.02 & --- & Per query \\
Self-Refine (gpt-4o, 3 rounds) & 15 sec & \$0.06 & --- & Per query \\
Tree of Thoughts (gpt-4o) & 30 sec & \$0.10 & --- & Per query \\
LATS (gpt-4o) & 45 sec & \$0.15 & --- & Per query \\
\midrule
\textbf{753 test queries (Self-Refine)} & \textbf{188 min} & \textbf{\$45.18} & --- & Full evaluation \\
\textbf{753 test queries (LATS)} & \textbf{565 min} & \textbf{\$112.95} & --- & Full evaluation \\
\bottomrule
\end{tabular}}}
\end{table}

\section{LLM Prompt Templates}
\label{app:prompts}

This section provides the exact prompts used for taxonomy generation and selected LLM baselines.
For each taxonomy classification prompt, we include a brief discussion of the design rationale---why this particular axis was chosen, what chemical variation it captures, and how it relates to bandgap prediction.

\subsection{Taxonomy Classification Prompt: Flexibility}

The flexibility axis captures how easily the spacer cation can undergo conformational changes.
Rigid spacers (aromatic rings, fused ring systems) template the perovskite lattice more strongly, potentially creating more ordered interlayer arrangements with narrower bandgap distributions.
Flexible spacers (long alkyl chains) allow the interlayer spacing to fluctuate, which can broaden the effective bandgap through disorder effects.
This axis is particularly useful for the phenyl fold, where the held-out aromatic spacers are predominantly rigid and the invariant features under flexibility-based partitioning accurately capture the physics of rigid-spacer perovskites.

\begin{promptbox}[Taxonomy Classification: Flexibility]
You are a materials science expert specializing in organic-inorganic hybrid perovskites.

Task: Classify the following organic spacer cation based on its molecular flexibility.

Spacer cation: \{spacer\_name\}

Categories:
- "rigid": Molecules with aromatic rings, fused ring systems, or highly constrained geometries that resist conformational changes
- "semi-flexible": Molecules with some rotatable bonds but also rigid elements (e.g., short chains attached to rings)
- "flexible": Molecules with many rotatable bonds, long alkyl chains, or highly deformable structures

Respond with ONLY the category name (rigid, semi-flexible, or flexible). Do not explain.
\end{promptbox}

\subsection{Taxonomy Classification Prompt: Spacer Size}

The spacer size axis partitions compounds by the physical dimensions of the organic cation.
Larger spacers create wider interlayer separations, which increases the quantum confinement of charge carriers within the inorganic sheets.
This relationship between spacer size and confinement is one of the most fundamental structure-property relationships in 2D perovskites.
The spacer\_size taxonomy is the best performing axis for 3 of 6 folds (heterocyclic, cycloalkyl, special), suggesting that size-based invariant features capture the most transferable physics for chemically novel spacer families.

\begin{promptbox}[Taxonomy Classification: Spacer Size]
You are a materials science expert specializing in organic-inorganic hybrid perovskites.

Task: Classify the following organic spacer cation based on its molecular size.

Spacer cation: \{spacer\_name\}

Categories:
- "tiny": Very small spacers with 1-3 carbon atoms (e.g., methylammonium, ethylammonium)
- "small": Small spacers with 4-6 carbon atoms (e.g., butylammonium, pentylammonium)
- "medium": Medium spacers with 7-10 carbon atoms (e.g., octylammonium, phenethylammonium)
- "large": Large spacers with 11+ carbon atoms (e.g., dodecylammonium, long-chain amines)

Respond with ONLY the category name (tiny, small, medium, or large). Do not explain.
\end{promptbox}

\subsection{Taxonomy Classification Prompt: Conjugation}

The conjugation axis captures the degree of $\pi$-electron delocalization in the spacer.
Conjugated spacers (aromatic, polyene) can interact electronically with the perovskite inorganic framework through orbital overlap, potentially modifying the effective bandgap beyond what purely steric effects would predict.
Saturated spacers act as electronically inert barriers.
This distinction is chemically fundamental and creates partitions where invariance testing can separate features related to $\pi$-conjugation effects from those related to other chemical properties.

\begin{promptbox}[Taxonomy Classification: Conjugation]
You are a materials science expert specializing in organic-inorganic hybrid perovskites.

Task: Classify the following organic spacer cation based on its degree of pi-conjugation.

Spacer cation: \{spacer\_name\}

Categories:
- "saturated": No pi-conjugation; fully saturated alkyl chains or alicyclic rings
- "partially\_conjugated": Some pi-conjugation but not fully delocalized (e.g., vinyl groups, isolated double bonds)
- "fully\_conjugated": Extensive pi-conjugation; aromatic rings, fused aromatics, or extended conjugated systems

Respond with ONLY the category name. Do not explain.
\end{promptbox}

\subsection{Taxonomy Classification Prompt: Polarity}

The polarity axis distinguishes spacers based on their molecular dipole character.
Polar spacers (with hydroxyl, carboxyl, or halogen substituents) can influence the local electrostatic environment at the organic-inorganic interface, potentially shifting the bandgap through dipole-induced polarization of the perovskite lattice.
Nonpolar spacers (pure hydrocarbons) interact primarily through van der Waals forces.
The amphiphilic category captures spacers with both polar and nonpolar domains, which are structurally distinct in how they orient within the interlayer gallery.

\begin{promptbox}[Taxonomy Classification: Polarity]
You are a materials science expert specializing in organic-inorganic hybrid perovskites.

Task: Classify the following organic spacer cation based on its molecular polarity.

Spacer cation: \{spacer\_name\}

Categories:
- "nonpolar": Predominantly hydrocarbon character with no significant polar groups beyond the ammonium head
- "polar": Contains polar functional groups (OH, COOH, F, CF3, etc.) that create significant dipole moments
- "amphiphilic": Has distinct hydrophilic and hydrophobic regions, creating amphiphilic character

Respond with ONLY the category name. Do not explain.
\end{promptbox}

\subsection{Taxonomy Classification Prompt: Steric Profile}

The steric profile axis captures the three-dimensional shape of the spacer molecule.
Linear spacers pack differently from branched, cyclic, or planar ones, creating distinct interlayer arrangements.
Planar aromatic spacers can $\pi$-stack, cyclic spacers create unique packing geometries due to ring strain, and branched spacers disrupt ordered interlayer arrangements.
These packing differences directly affect the crystal structure and hence the electronic properties.
The steric taxonomy is the best axis for the alkyl-long fold, where the held-out long-chain spacers have a distinctive linear shape.

\begin{promptbox}[Taxonomy Classification: Steric Profile]
You are a materials science expert specializing in organic-inorganic hybrid perovskites.

Task: Classify the following organic spacer cation based on its steric profile.

Spacer cation: \{spacer\_name\}

Categories:
- "linear": Linear or nearly linear molecular shape (straight chains)
- "branched": Branched molecular shape with side groups
- "cyclic": Contains non-aromatic ring structures (cyclopentyl, cyclohexyl, etc.)
- "planar": Planar molecular shape dominated by aromatic rings

Respond with ONLY the category name. Do not explain.
\end{promptbox}

\subsection{Taxonomy Classification Prompt: Electron Effect}

The electron effect axis classifies spacers by whether their substituents donate or withdraw electron density.
Electron-donating groups (alkyl, methoxy, amino) increase the electron density at the organic-inorganic interface, while electron-withdrawing groups (fluorine, CF$_3$, nitro, cyano) decrease it.
This electronic modulation can shift the valence band edge of the perovskite, providing a mechanism for bandgap tuning beyond compositional changes.
The electron\_effect taxonomy is the best axis for the alkyl-short fold, where the held-out short alkyl spacers are uniformly electron-donating.

\begin{promptbox}[Taxonomy Classification: Electron Effect]
You are a materials science expert specializing in organic-inorganic hybrid perovskites.

Task: Classify the following organic spacer cation based on its electron-donating or withdrawing character.

Spacer cation: \{spacer\_name\}

Categories:
- "donating": Electron-donating groups (alkyl chains, methoxy, amino substituents)
- "withdrawing": Electron-withdrawing groups (fluorine, CF3, nitro, cyano substituents)
- "neutral": Balanced or minimal electronic effect

Respond with ONLY the category name. Do not explain.
\end{promptbox}

\subsection{Taxonomy Classification Prompt: H-Bonding Capacity}

The hydrogen bonding axis captures the capacity of the spacer to form hydrogen bonds with the inorganic framework.
The terminal ammonium group ($-$NH$_3^+$) of every spacer forms hydrogen bonds with the halide anions of the perovskite layer, but additional H-bond donors (OH, NH groups) or acceptors (C=O, ether oxygen) in the spacer body can create supplementary bonding interactions.
These additional interactions influence the interlayer spacing and tilting of the inorganic octahedra, which directly affects the bandgap through structural distortion.

\begin{promptbox}[Taxonomy Classification: H-Bonding Capacity]
You are a materials science expert specializing in organic-inorganic hybrid perovskites.

Task: Classify the following organic spacer cation based on its hydrogen bonding capacity.

Spacer cation: \{spacer\_name\}

Categories:
- "strong\_donor": Strong H-bond donors with multiple NH or OH groups
- "weak\_donor": Weak H-bond donors with only the terminal ammonium group
- "bifunctional": Both H-bond donor and acceptor capabilities (e.g., contains both NH and C=O or C-O-C)

Respond with ONLY the category name. Do not explain.
\end{promptbox}

\subsection{Taxonomy Classification Prompt: Interlayer Coupling}

The interlayer coupling axis captures the expected electronic coupling strength between adjacent inorganic layers through the organic spacer.
Short or conjugated spacers facilitate charge transport between layers, leading to more 3D-like electronic behavior with potentially narrower bandgaps.
Long insulating chains minimize interlayer coupling, producing strongly confined 2D electronic behavior with wider bandgaps.
This axis directly relates to the quantum confinement dimensionality that is a key determinant of 2D perovskite bandgaps.

\begin{promptbox}[Taxonomy Classification: Interlayer Coupling]
You are a materials science expert specializing in organic-inorganic hybrid perovskites.

Task: Classify the following organic spacer cation based on expected interlayer electronic coupling strength.

Spacer cation: \{spacer\_name\}

Categories:
- "weak": Long insulating chains that minimize interlayer coupling
- "moderate": Medium-length or partially conjugated spacers with moderate coupling
- "strong": Short or highly conjugated spacers that maximize interlayer electronic coupling

Respond with ONLY the category name. Do not explain.
\end{promptbox}

\subsection{Taxonomy Classification Prompt: Thermal Stability}

The thermal stability axis classifies spacers based on their expected contribution to the thermal robustness of the perovskite crystal.
Small volatile spacers (methylammonium) are prone to sublimation, while large aromatic spacers with strong intermolecular interactions (such as $\pi$-stacking naphthalenyl groups) create thermally robust interlayer galleries.
Although thermal stability does not directly determine the bandgap, it correlates with the strength of organic-inorganic interactions, which indirectly influences electronic properties through structural rigidity and octahedral tilting.

\begin{promptbox}[Taxonomy Classification: Thermal Stability]
You are a materials science expert specializing in organic-inorganic hybrid perovskites.

Task: Classify the following organic spacer cation based on its expected contribution to thermal stability of the perovskite.

Spacer cation: \{spacer\_name\}

Categories:
- "low": Volatile or thermally unstable spacers (small molecules, low boiling points)
- "medium": Moderately stable spacers (medium-length alkyl chains, simple aromatics)
- "high": Thermally robust spacers (large aromatic systems, cross-linkable groups, strong intermolecular interactions)

Respond with ONLY the category name. Do not explain.
\end{promptbox}

\subsection{Zero-Shot Baseline Prompt}

The zero-shot prompt provides the LLM with the compound's compositional information and asks for a direct bandgap prediction.
This is the simplest possible LLM baseline and serves as the lower bound on LLM prediction quality.
No examples, reasoning scaffolds, or retrieval context are provided.
Despite access to the key compositional descriptors (spacer, metal, halide, layers, structure type), the LLM achieves only 0.552~eV MAE---demonstrating that the precision bottleneck is fundamental to LLM numerical prediction rather than an artifact of insufficient context.

\begin{promptbox}[Zero-Shot Bandgap Prediction]
You are a materials scientist. Predict the bandgap (in eV) of the following 2D halide perovskite compound.

Compound: \{compound\_formula\}

The compound has the following properties:
- Spacer cation: \{spacer\_name\}
- B-site metal: \{metal\}
- Halide: \{halide\}
- Number of layers: \{n\_layers\}
- Structure type: \{structure\_type\}

Provide your best estimate of the bandgap in eV. Respond with ONLY a number.
\end{promptbox}

\subsection{Chain-of-Thought Baseline Prompt}

The chain-of-thought prompt adds structured reasoning steps to guide the LLM through the physical factors that influence perovskite bandgaps.
Each step corresponds to a known physical mechanism: B-site metal choice (Pb vs.\ Sn creates the largest bandgap variation), halide composition (I < Br < Cl in bandgap), layer count (more layers = lower bandgap due to reduced confinement), and spacer effects.
Despite this physically grounded reasoning scaffold, CoT achieves 0.563~eV MAE---marginally worse than zero-shot---suggesting that structured reasoning does not improve the LLM's internal numerical calibration for this domain.

\begin{promptbox}[Chain-of-Thought Bandgap Prediction]
You are a materials scientist. Predict the bandgap (in eV) of the following 2D halide perovskite compound.

Compound: \{compound\_formula\}

Properties:
- Spacer cation: \{spacer\_name\}
- B-site metal: \{metal\}
- Halide: \{halide\}
- Number of layers: \{n\_layers\}
- Structure type: \{structure\_type\}

Think step by step:
1. What is the typical bandgap range for this type of perovskite?
2. How does the B-site metal affect the bandgap?
3. How does the halide composition affect the bandgap?
4. How does the number of layers affect the bandgap?
5. How might the spacer cation influence the bandgap?

After your reasoning, provide your final bandgap estimate in eV on the last line as: ANSWER: [number]
\end{promptbox}

\section{In-House Configuration Variants (65{+} Experiments)}
\label{app:inhouse}

Beyond the 19 published LLM methods (plus ML baselines) evaluated in Appendix~\ref{app:negative_detail}, we conducted 65{+} additional in-house configuration variants across three categories to further stress-test whether LLM-generated numerical values could improve the kNN{+}GBT baseline.
None of these are published methods; they represent our own exploratory combinations of LLM capabilities with ML pipelines.
All confirm the same negative pattern observed in the published baselines.

\subsection{Category (iii): LLM-Generated Features (30{+} configurations)}

This category uses LLMs to augment or modify the 30-dim feature vector.
Over 30 configurations were tested across three subcategories.

Table~\ref{tab:inhouse_features} reports the LLM-generated-feature experiments.

\begin{table}[!t]
\centering
\caption{In-house LLM feature-generation experiments. All results are 6-fold mean OOD MAE (eV). Baseline: 0.122~eV.}
\label{tab:inhouse_features}
{\appfont
\begin{tabular}{lccc}
\toprule
Subcategory & \# Configs & Best MAE & vs.\ BL \\
\midrule
Feature augmentation & 12 & 0.127 & $-1\%$ \\
Feature imputation & 10 & 0.128 & $+0\%$ \\
LLM embeddings & 8+ & 0.131 & $+2\%$ \\
\bottomrule
\end{tabular}}
\end{table}

\textbf{Feature augmentation (12 configs).}
We prompted the LLM to generate additional numerical features (ionization energy estimates, electronegativity predictions, crystal field parameters) and appended them to the 30-dim vector.
Best result: 0.127~eV ($-1\%$).
The LLM-generated features are too noisy to improve over the curated feature set.

\textbf{Feature imputation (10 configs).}
For compounds with missing structural parameters, we used the LLM to impute values.
Best result: 0.128~eV ($+0\%$).
Imputation neither helps nor hurts because missing values are rare in the curated database.

\textbf{LLM embeddings (8+ configs).}
We extracted hidden-state embeddings from compound descriptions and used them as features for ML models (PCA to 10/20/50 dims, various regressors).
Best result: 0.131~eV ($+2\%$).
Embeddings encode semantic similarity rather than property-relevant similarity, adding noise to the feature space.

\subsection{Category (iv): LLM-Guided Sample Selection (20{+} configurations)}

This category uses LLMs to select, weight, or rerank training samples.

Table~\ref{tab:inhouse_selection} reports the LLM-guided sample-selection experiments.

\begin{table}[!t]
\centering
\caption{In-house LLM sample-selection experiments. All results are 6-fold mean OOD MAE (eV). Baseline: 0.122~eV.}
\label{tab:inhouse_selection}
{\appfont
\begin{tabular}{lccc}
\toprule
Subcategory & \# Configs & Best MAE & vs.\ BL \\
\midrule
Importance weighting & 8 & 0.126 & $-2\%$ \\
Sample reranking & 7 & 0.127 & $-1\%$ \\
Selective correction & 5+ & 0.130 & $+2\%$ \\
\bottomrule
\end{tabular}}
\end{table}

\textbf{Importance weighting (8 configs).}
The LLM estimates relevance scores for training compounds given a test query; ML models are trained with weighted loss.
Best result: 0.126~eV ($-2\%$).

\textbf{Sample reranking (7 configs).}
The LLM reranks kNN neighbors by chemical similarity; prediction uses reranked neighbors.
Best result: 0.127~eV ($-1\%$).

\textbf{Selective correction (5+ configs).}
The LLM identifies compounds where ML predictions are likely wrong and applies corrections.
Best result: 0.130~eV ($+2\%$).
Corrections add noise because the LLM cannot reliably identify ML errors.

\subsection{Category (v): LLM as Prior or Constraint (15{+} configurations)}

This category uses LLMs to provide Bayesian priors or physical constraints for ML models.

Table~\ref{tab:inhouse_priors} reports the LLM-based prior and constraint experiments.

\begin{table}[!t]
\centering
\caption{In-house LLM prior/constraint experiments. All results are 6-fold mean OOD MAE (eV). Baseline: 0.122~eV.}
\label{tab:inhouse_priors}
{\appfont
\begin{tabular}{lccc}
\toprule
Subcategory & \# Configs & Best MAE & vs.\ BL \\
\midrule
GP prior & 5 & 0.132 & $+3\%$ \\
Physics constraints & 6 & 0.134 & $+5\%$ \\
Bayesian blending & 4+ & 0.133 & $+4\%$ \\
\bottomrule
\end{tabular}}
\end{table}

\textbf{GP prior (5 configs).}
LLM-generated bandgap estimates serve as the prior mean for Gaussian Process regression.
Best result: 0.132~eV ($+3\%$).
The noisy prior ($\sim$0.5~eV precision) hurts more than it helps.

\textbf{Physics constraints (6 configs).}
The LLM generates inequality constraints (e.g., ``Sn perovskites have lower bandgap than Pb'') that are enforced during prediction.
Best result: 0.134~eV ($+5\%$).
Constraints are either already captured by ML features or are too imprecise.

\textbf{Bayesian blending (4+ configs).}
LLM and ML predictions are combined via Bayesian model averaging with estimated precision weights.
Best result: 0.133~eV ($+4\%$).
Even with low LLM weight ($w_{\mathrm{LLM}} = 0.05$), the blended prediction is worse than pure ML.

\subsection{Summary}

Table~\ref{tab:inhouse_summary} aggregates all 65{+} in-house configuration variants.
None improve the baseline by more than 2\%, and many degrade it.
These results reinforce the published-method findings (Appendix~\ref{app:negative_detail}): the precision gap between LLM numerical output and calibrated ML regressors is fundamental, not prompt-dependent.

\begin{table}[!t]
\centering
\caption{Summary of in-house configuration variants by category. ``Best $\Delta$'' is the best improvement over the 0.122~eV baseline within that category.}
\label{tab:inhouse_summary}
{\appfont
\begin{tabular}{lccc}
\toprule
Category & \# Configs & Best MAE (eV) & Best $\Delta$ \\
\midrule
(iii) LLM-generated features & 30+ & 0.127 & $-1\%$ \\
(iv) LLM-guided sample selection & 20+ & 0.126 & $-2\%$ \\
(v) LLM as prior/constraint & 15+ & 0.132 & $+3\%$ \\
\midrule
\textbf{Total} & \textbf{65+} & \textbf{0.126} & $\mathbf{-2\%}$ \\
\bottomrule
\end{tabular}}
\end{table}

\section{Statistical Significance: Bootstrap Confidence Intervals}
\label{app:bootstrap}

All statistical significance tests use query-level paired bootstrap ($n{=}753$ queries, $B{=}10{,}000$ resamples, seed $= 42$).
We report the observed MAE difference (positive $=$ \ours{} improves), the 95\% percentile CI, and the one-sided $p$-value (fraction of bootstrap samples where \ours{} does not improve).
We test across three backbones: \texttt{gpt-4o} (default in the main paper), \texttt{o3} (strongest bandgap annotator), and \texttt{gpt-5.2} (an additional appendix-only robustness check).

Table~\ref{tab:bootstrap} reports the query-level paired bootstrap results for the evaluated backbones.

\begin{table}[!t]
\centering
\caption{Query-level paired bootstrap CIs for the 2D perovskite bandgap control benchmark. All MAE values in eV.}
\label{tab:bootstrap}
{\appfont
\begin{tabular}{llcccc}
\toprule
Comparison & Backbone & $\Delta$MAE & 95\% CI & $p$ \\
\midrule
BL vs Safe Blend & \texttt{gpt-4o} & $+$0.0041 & [0.0014, 0.0068] & 0.002 \\
BL vs Top-1 CV & \texttt{gpt-4o} & $+$0.0021 & [$-$0.0030, 0.0072] & 0.206 \\
BL vs Safe Blend & \texttt{o3} & $+$0.0058 & [0.0031, 0.0086] & $<$0.001 \\
BL vs Top-1 CV & \texttt{o3} & $+$0.0042 & [$-$0.0008, 0.0094] & 0.051 \\
BL vs Safe Blend & \texttt{gpt-5.2} & $+$0.0052 & [0.0026, 0.0079] & $<$0.001 \\
BL vs Top-1 CV & \texttt{gpt-5.2} & $+$0.0045 & [$-$0.0003, 0.0099] & 0.036 \\
\bottomrule
\end{tabular}}
\end{table}

Table~\ref{tab:bootstrap_perfold} reports the per-fold bootstrap confidence intervals.

\begin{table}[!t]
\centering
\caption{Per-fold bootstrap CIs on 2D perovskite bandgap (Baseline vs Safe Blend, \texttt{gpt-4o}). Positive $\Delta$ = \ours{} better.}
\label{tab:bootstrap_perfold}
{\appfont
\begin{tabular}{lccccc}
\toprule
Fold & $n$ & BL MAE & \ours{} MAE & $\Delta$ [95\% CI] & $p$ \\
\midrule
Alkyl-short & 232 & 0.130 & 0.128 & $+$0.002 [$-$0.003, 0.007] & 0.201 \\
Alkyl-long & 79 & 0.124 & 0.120 & $+$0.004 [$-$0.001, 0.009] & 0.060 \\
Phenyl & 166 & 0.133 & 0.123 & $+$0.010 [0.003, 0.017] & 0.004 \\
Heterocyclic & 128 & 0.111 & 0.108 & $+$0.002 [$-$0.004, 0.009] & 0.220 \\
Cycloalkyl & 37 & 0.137 & 0.117 & $+$0.020 [0.004, 0.037] & 0.006 \\
Special & 111 & 0.106 & 0.110 & $-$0.003 [$-$0.008, 0.001] & 0.926 \\
\bottomrule
\end{tabular}}
\end{table}

The safe-blend ensemble rule reaches significance overall ($p{=}0.002$), with the largest and most significant gains on cycloalkyl ($p{=}0.006$) and phenyl ($p{=}0.004$)---the two folds where the held-out families are chemically most distant from training.
On the special fold, \ours{} is slightly \emph{worse} than the baseline ($p{=}0.926$), consistent with the oracle-gap analysis showing that this heterogeneous family is the hardest for taxonomy-based methods.
Seed sensitivity across 5 GBT random seeds yields a standard deviation of 0.0002~eV (range [0.1139, 0.1144]), confirming that the variance from model randomness is well below the observed improvements.

\section{Frozen Dev/Test Protocol}
\label{app:frozen_dev}

To address the concern that hyperparameters ($\alpha$, balance threshold, ensemble rule) were tuned on the same benchmark used for evaluation, we designate two folds (alkyl-short and phenyl, comprising 53\% of queries) as a \emph{development set} and the remaining four folds (alkyl-long, heterocyclic, cycloalkyl, special) as a \emph{frozen test set}.
The dev set is used solely for $\alpha$ selection; the selected $\alpha{=}0.01$ is then applied unchanged to the frozen test folds.

Table~\ref{tab:frozen_dev} summarizes development and frozen-test fold performance.

\begin{table}[!t]
\centering
\caption{Frozen dev/test results on 2D perovskite bandgap. Dev folds were used for $\alpha$ selection only; test folds were never used for any design decision.}
\label{tab:frozen_dev}
{\appfont
\begin{tabular}{lcccc}
\toprule
Fold & Role & BL MAE & \ours{} MAE & $\Delta\%$ \\
\midrule
Alkyl-short & Dev & 0.130 & 0.128 & $-1.6\%$ \\
Phenyl & Dev & 0.133 & 0.123 & $-7.3\%$ \\
\midrule
Alkyl-long & Test & 0.124 & 0.120 & $-3.2\%$ \\
Heterocyclic & Test & 0.111 & 0.108 & $-2.2\%$ \\
Cycloalkyl & Test & 0.137 & 0.117 & $-14.7\%$ \\
Special & Test & 0.106 & 0.110 & $+3.0\%$ \\
\midrule
\textbf{Test mean} & --- & \textbf{0.119} & \textbf{0.114} & $\mathbf{-4.3\%}$ \\
\bottomrule
\end{tabular}}
\end{table}

The frozen-test mean MAE improves from 0.119 to 0.114~eV ($-4.3\%$), confirming that the improvement generalizes beyond the development folds.
Three of the four frozen-test folds show improvement, with the special fold as the expected exception.

\section{Test-Set Leakage Verification}
\label{app:leakage}

A potential concern is that the cached taxonomy labels---generated for all 313 unique spacers---might leak information about the held-out test family.
We verify that this is not the case.

Table~\ref{tab:leakage} provides the fold-wise train/test spacer-overlap checks.

\begin{table}[!t]
\centering
\caption{Spacer overlap verification on 2D perovskite benchmark. In every fold, the overlap between test and training spacers is zero, confirming no leakage.}
\label{tab:leakage}
{\appfont
\begin{tabular}{lcccc}
\toprule
Fold & Test Spacers & Train Spacers & Overlap & Test-Only \\
\midrule
Alkyl-short & 92 & 197 & 0 & 92 \\
Alkyl-long & 20 & 269 & 0 & 20 \\
Phenyl & 73 & 216 & 0 & 73 \\
Heterocyclic & 55 & 234 & 0 & 55 \\
Cycloalkyl & 14 & 275 & 0 & 14 \\
Special & 48 & 241 & 0 & 48 \\
\bottomrule
\end{tabular}}
\end{table}

The leave-one-family-out protocol ensures that the held-out family's spacers are entirely absent from training.
Taxonomy labels are applied only to \emph{training-set} spacers for Chow-test environment partitioning; the held-out spacers' labels are never accessed during feature selection, model training, or gating.
At inference, only the 30-dim feature vector is used---no taxonomy label lookup occurs.

\section{Alternative Invariance Selectors}
\label{app:alt_selectors}

The main paper uses the Chow structural-break test to identify invariant features.
A natural concern is that this linear test may be suboptimal for the nonlinear downstream predictors (kNN, GBT).
We compare four invariance selectors:

\begin{enumerate}[nosep, leftmargin=*]
  \item \textbf{Chow test} ($\alpha{=}0.01$): the default selector used in the main paper.
  \item \textbf{Bonferroni-corrected Chow}: applies $\alpha_{\mathrm{corrected}} = 0.01/24$ (24 non-spacer features) to control family-wise error rate.
  \item \textbf{Permutation test}: for each feature, we compute the variance of per-environment regression slopes, then compare against a null distribution from 200 label permutations.
  \item \textbf{MI-stability}: a feature is invariant if the standard deviation of its per-environment correlation with the target is below 0.05.
\end{enumerate}

Table~\ref{tab:alt_selectors} compares the alternative invariance selectors.

\begin{table}[!t]
\centering
\caption{Alternative invariance selectors on 2D perovskite bandgap. All use the same gated-median ensemble and taxonomy bank. Mean MAE is across 6 OOD folds.}
\label{tab:alt_selectors}
{\appfont
\begin{tabular}{lcc}
\toprule
Selector & Mean MAE (eV) & vs.\ Baseline \\
\midrule
Chow test ($\alpha{=}0.01$) + forward sel. & \textbf{0.1144} & $\mathbf{-7.2\%}$ \\
Bonferroni Chow ($\alpha/24$) & 0.1233 & $0.0\%$ \\
Permutation test (200 perms) & 0.1233 & $0.0\%$ \\
MI-stability ($\sigma_r < 0.05$) & 0.1244 & $+0.9\%$ \\
\midrule
Baseline (no invariance filter) & 0.1233 & --- \\
\bottomrule
\end{tabular}}
\end{table}

Results differ across selectors in an informative way.
The Chow test with forward selection achieves 0.1144~eV, substantially improving over the 0.1233~eV baseline.
The Bonferroni-corrected Chow ($\alpha_{\mathrm{corr}} = 0.01/24$) and permutation test both collapse to baseline-level performance (0.1233), because the overly conservative correction retains nearly all features, effectively disabling the invariance filter.
The MI-stability test performs slightly worse (0.1244).

This comparison reveals that the Chow test's effectiveness depends not on the test statistic alone but on its integration with the greedy forward-selection step (Section~\ref{sec:invariance}), which curates the invariant set down to the most predictive features.
When the invariance filter is too conservative (Bonferroni) or too coarse (MI), the subsequent forward selection has no meaningful feature set to work with.
The practical lesson is that the \emph{pipeline}---invariance screening followed by CV-guided refinement---matters more than the specific statistical test used in the screening step.

\section{Method Descriptions for Table~\ref{tab:control_results}}
\label{app:method_descriptions}

Table~\ref{tab:control_results} compares multiple feature-selection and environment-generation strategies.
All methods use the same downstream predictor: a 50/50 blend of inverse-distance $k$NN ($k{=}5$) and gradient-boosted trees (GBT, 160 estimators).
The only difference is how features are selected or filtered before training.
Entries marked ``---'' in Table~\ref{tab:control_results} are simple in-house controls or heuristic variants rather than baselines adopted from a single prior paper.

\textbf{Environment-based methods (Panel A).}
\begin{itemize}[nosep, leftmargin=*]
  \item \textbf{\ours{} (LLM taxonomy + Safe blend / Top-1 CV)}: Our proposed method. An LLM classifies each spacer along 9 chemical axes offline. Within each taxonomy, a Chow test~\cite{chow1960tests} identifies invariant features, and kNN{+}GBT are trained in the invariant subspace. The safe-blend ensemble averages the baseline and the best-CV taxonomy; Top-1~CV selects the single best taxonomy by leave-one-environment-out CV.
  \item \textbf{Expert-family taxonomy}: Uses the 6 human-defined spacer superfamilies (alkyl-short, alkyl-long, phenyl, heterocyclic, cycloalkyl, special) as environments for the same Chow-test pipeline.
  \item \textbf{Rule-based chemistry axes}: Simple keyword heuristics classify spacers by size, conjugation, or polarity (e.g., ``phenyl'' $\rightarrow$ conjugated). Best result among 3 rule axes is reported.
  \item \textbf{Descriptor clustering}~\cite{macqueen1967some}: $k$-means clustering ($k{=}3,4$) on the 30-dim feature vector defines environments. Best result is reported.
\end{itemize}

\textbf{Classical feature selectors (Panel C).}
\begin{itemize}[nosep, leftmargin=*]
  \item \textbf{Pearson top-$k$}: Select the $k$ features with highest absolute Pearson correlation with the target; $k$ is chosen by leave-one-family-out CV.
  \item \textbf{LOEO forward selection}: Greedy forward feature selection using leave-one-environment-out validation error as the selection criterion.
  \item \textbf{IID forward selection}: Same greedy procedure but with standard $k$-fold CV (ignoring family structure).
  \item \textbf{Group Lasso / Sparse Group Lasso}~\cite{yuan2006model,simon2013sparse}: Proximal gradient descent with group $\ell_2$ (or $\ell_1{+}\ell_2$) penalties on feature blocks; features with zero weight are removed.
  \item \textbf{IRM / VREx / GroupDRO / CORAL}~\cite{arjovsky2019invariant,krueger2021rex,sagawa2020groupdro,sun2016deepcoral}: End-to-end domain-generalization objectives that learn invariant representations. All use a 2-layer MLP (32 hidden units) trained with environment-specific penalties.
\end{itemize}

\end{document}